\documentclass[10pt,twocolumn,letterpaper]{article}

\usepackage{iccv}
\usepackage[utf8]{inputenc}
\usepackage{bm}
\usepackage{times}
\usepackage{epsfig}
\usepackage{enumitem}
\usepackage{graphicx}
\usepackage{amsmath}
\usepackage{amssymb}
\usepackage{algorithm}
\usepackage{algpseudocode}
\usepackage{latexsym}
\usepackage{mathtools}
\usepackage{multirow}
\usepackage{nicefrac}
\usepackage{makecell}
\usepackage{rotating}
\usepackage{url}
\usepackage{xspace}
\usepackage{wrapfig}
\usepackage{subfig}
\usepackage{microtype}
\usepackage{booktabs}
\usepackage[dvipsnames]{xcolor}
\usepackage[colorlinks,breaklinks=true,bookmarks=false,pagebackref,letterpaper]{hyperref}
\hypersetup{
    pdfauthor={Jérôme Rony, Eric Granger, Marco Pedersoli, Ismail {Ben Ayed}},
    pdftitle=Augmented Lagrangian Adversarial Attacks,
    linkcolor=BrickRed,
    citecolor=Green,
}
\DeclareMathAlphabet{\mathalphm}{OMS}{cmsy}{m}{n}
\DeclareMathAlphabet{\mathalphb}{OMS}{cmsy}{b}{n}

\newcommand{\x}{\bm{x}}
\newcommand{\xadv}{\tilde{\x}}
\newcommand{\ytrue}{y}

\newcommand{\g}{\bm{g}}

\newcommand{\z}{\bm{z}}
\newcommand{\N}{\mathbb{N}\xspace}

\newcommand{\Y}{\mathcal{Y}}
\newcommand{\X}{\mathcal{X}}

\newcommand{\Z}{\mathalphm{Z}}

\DeclareMathOperator{\softmax}{softmax}

\makeatletter
\DeclareRobustCommand\onedot{\futurelet\@let@token\@onedot}
\def\@onedot{\ifx\@let@token.\else.\null\fi\xspace}

\def\eg{\emph{e.g}\onedot} 
\def\ie{\emph{i.e}\onedot}

\def\wrt{w.r.t\onedot} 
\def\etal{\emph{et al}\onedot}
\makeatother

\newcommand{\R}{\mathbb{R}}

\DeclareMathOperator*{\argmax}{arg\,max}
\makeatletter

\newcommand\Autoref[1]{\@first@ref#1,@}
\def\@throw@dot#1.#2@{#1}
\def\@set@refname#1{
    \edef\@tmp{\getrefbykeydefault{#1}{anchor}{}}%
    \xdef\@tmp{\expandafter\@throw@dot\@tmp.@}%
    \ltx@IfUndefined{\@tmp autorefnameplural}%
         {\def\@refname{\@nameuse{\@tmp autorefname}s}}%
         {\def\@refname{\@nameuse{\@tmp autorefnameplural}}}%
}
\def\@first@ref#1,#2{%
  \ifx#2@\autoref{#1}\let\@nextref\@gobble
  \else%
    \@set@refname{#1}
    \@refname~\ref{#1}
    \let\@nextref\@next@ref
  \fi%
  \@nextref#2%
}
\def\@next@ref#1,#2{%
   \ifx#2@ and~\ref{#1}\let\@nextref\@gobble
   \else, \ref{#1}
   \fi%
   \@nextref#2%
}

\makeatother

\setlist{nosep}  

\iccvfinalcopy 

\def\iccvPaperID{6379} 

\ificcvfinal\fi

\begin{document}

\title{\vspace{-2em}Augmented Lagrangian Adversarial Attacks}

\author{
    Jérôme Rony\thanks{Corresponding author: \url{jerome.rony.1@etsmtl.net}} 
    \qquad Eric Granger
    \qquad Marco Pedersoli
    \qquad Ismail {Ben Ayed}\\
    ÉTS Montreal, Canada\\
    {\small Code: \url{https://github.com/jeromerony/augmented_lagrangian_adversarial_attacks}}
}

\maketitle
\ificcvfinal\thispagestyle{empty}\fi

\begin{abstract}

Adversarial attack algorithms are dominated by penalty methods, which are slow in practice, or more efficient distance-customized methods, which are heavily tailored to the properties of the distance considered. We propose a white-box attack algorithm to generate minimally perturbed adversarial examples based on Augmented Lagrangian principles. We bring several algorithmic modifications, which have a crucial effect on performance. Our attack enjoys the generality of penalty methods and the computational efficiency of distance-customized algorithms, and can be readily used for a wide set of distances. We compare our attack to state-of-the-art methods on three datasets and several models, and consistently obtain competitive performances with similar or lower computational complexity.

\end{abstract}

\section{Introduction}

The last few years have seen an arms race in adversarial attacks, where several methods have been proposed to find minimally perturbed adversarial examples. 
In most cases, adversarial example generation is stated as a constrained optimization, which seeks the smallest additive perturbation, according to some distance, to misclassify an input. Existing methods fall within two categories: attacks using {\em penalty} methods for constrained optimization, and distance-customized attacks leveraging the properties of the distance considered (typically $\ell_p$-norms). 

Penalties are a natural choice, as they transform a constrained-optimization problem into an unconstrained one. Within this category, the most notorious attack is the one from Carlini and Wagner \cite{carlini2017towards} known for its $\ell_2$ variant. Building on the work of Carlini and Wagner, several other attacks have been proposed, \eg EAD for $\ell_1$ \cite{chen2018ead} and  StrAttack \cite{xu2018structured}, which generalizes EAD and introduces sparsity in the perturbations. 
More recently, other works have tackled other distances than the standard $\ell_p$-norms, such as SSIM \cite{gragnaniello2019perceptual}, CIEDE2000 \cite{zhao2020towards} or LPIPS \cite{laidlaw2021perceptual}, and followed similar penalty-based strategies. Although  convenient and applicable to a wide class of distances, penalty methods are known to result in slow convergence in the general field of optimization \cite{jensen2003operations}. Furthermore, choosing the weight of the penalty is not a trivial task \cite{kervadec2019constraineddeep}. In fact, in adversarial attacks, it has recently been shown that the optimal penalty weight actually varies by orders of magnitude across samples and models \cite{rony2019decoupling}. Although penalty methods can achieve competitive performance, they typically require an expensive line-search to find optimal penalty weights \cite{carlini2017towards}, thereby requiring large numbers of iterations. This may impede their practical deployment for training robust models and efficiently evaluating robustness.

To accelerate the generation of adversarial examples, and improve performance over penalty methods, there has been an intensive focus on developing efficient algorithms customized for specific $\ell_p$-norms \cite{brendel2019accurate, croce2020minimally, moosavi2016deepfool, pintor2021fast, rony2019decoupling}. However, such distance-customized methods are not generally applicable because they rely heavily on the geometry/properties of the distance considered (\eg using projections and dual norms) to find minimally perturbed adversarial examples. The most notable attacks within this second category are: DeepFool for the $\ell_2$ and $\ell_\infty$ norms, which uses a linear approximation of the model at each iteration \cite{moosavi2016deepfool}; DDN for the $\ell_2$-norm, which uses projections on the $\ell_2$-ball to decouple the direction and the norm of the perturbation \cite{rony2019decoupling}; and FAB for the $\ell_1$, $\ell_2$ and $\ell_\infty$, which combines a linear approximation of the model and projections \wrt the norm considered \cite{croce2020minimally}. The recently proposed FMN attack \cite{pintor2021fast} extends the DDN attack to other norms. Beyond $\ell_p$-norms, Wong \etal proposed an attack \cite{wong2019wasserstein} to produce adversarial perturbations with minimal Wasserstein distance using projected Sinkhorn iterations. Finally, one attack that does not strictly fall in either of the two categories was proposed by Brendel \etal, and designed for $\ell_p$-norms with $p\in\{0, 1, 2, \infty\}$ \cite{brendel2019accurate}. In this attack, the optimization is formulated such that the perturbation follows the decision boundary of a classifier, while minimizing the considered distance. This is not limited to $\ell_p$-norms but, in practice, the implementation leverages a trust-region solver designed for each $\ell_p$-norm specifically, which limits applicability to other distances. 


Penalty methods are generally applicable, and can be used for distances other than the standard $\ell_p$-norms; for instance, CIEDE2000 \cite{zhao2020towards} or LPIPS \cite{laidlaw2021perceptual, zhang2018unreasonable}. They replace constrained problems with unconstrained ones by adding a penalty, which increases when the constraint is violated. A weight of the penalty is chosen and increased heuristically, while the unconstrained optimization is repeated several times. Powerful Augmented Lagrangian principles have well-established advantages over penalties in the general context of optimization \cite{bertsekas2014constrained, bertsekas2016nonlinear, fletcher2013practical, nocedal2006numerical}, and completely avoid penalty-weight heuristics by automatically estimating the multipliers. Furthermore, the multiplier estimates tend to the Lagrange multipliers, which avoids the ill-conditioning problems often encountered in penalty methods \cite{bertsekas2014constrained, conn1997globally}. Finally, Augmented Lagrangian methods avoid the need for explicitly solving the dual problem, unlike basic Lagrangian-dual optimization, which might be intractable/unstable for non-convex problems. 

Despite their well-established advantages and popularity in the optimization community for solving non-convex problems, Augmented Lagrangian methods have not been investigated previously for adversarial attacks. 
While this seems rather surprising, we found in this work that the vanilla Augmented Lagrangian methods are not competitive in the context of adversarial attacks (\eg in terms of computational efficiency), which might explain why they have been avoided so far. However, we introduce several algorithmic modifications to design a customized Augmented Lagrangian algorithm for adversarial example generation. Our modifications are crucial to achieve highly competitive performances in comparisons to state-of-the-art methods. The modifications include integrating the Augmented Lagrangian inner and outer iterations with joint updates of the perturbation and the multipliers, relaxing the need for an inner-convergence criterion, introducing an exponential moving average of the multipliers, and adapting the learning rates to the distance function. All-in-all, we propose a white-box Augmented Lagrangian Method for Adversarial (ALMA) attacks, which enjoys both the general applicability of penalty approaches and computational efficiency of distance-customized methods. Our attack can be readily used to generate adversarial examples for a large set of distances, including  $\ell_1$-norm, $\ell_2$-norm, CIEDE2000, LPIPS and SSIM, and we advocate its use for other distances that might be investigated in future research in adversarial attacks. We evaluate our attack on three datasets (MNIST, CIFAR10 and ImageNet) and several models (regularly and adversarially-trained). For each distance, we compare our method against state-of-the-art attacks proposed specifically for that distance, and consistently observe competitive performance.

\section{Preliminaries}

Let $\x$ be a sample from the input space $\X\subset\R^d$, and $\ytrue\in\Y$ its associated label, where $\Y$ is a set of discrete labels of size K.
Let $f:\R^d\rightarrow \R^K$ be a model that outputs logits (\ie pre-softmax scores) $\z\in\R^K$ given an input $\x$; $f_k(\x)$ denotes the $k$-th component of the vector $f(\x)$. In a classification scenario, the probability $p_y = P(y|x)$ is obtained using the $\softmax$ function: $p_y = \softmax_y(\z)$. 
In this work, we assume that $\X$ is the hypercube $\X=[0,1]^d$, which is general enough for computer vision applications.

\subsection{Problem formulation}
\label{sec:problem_formulation}

The problem of adversarial example generation has been mainly formulated in two ways. 
One way is to find adversarial examples satisfying a distance constraint: 
\begin{equation}
\begin{aligned}
\label{eq:budget_adv_optim}
&\text{find} & & \delta & & \text{s.t.} & & \argmax_k f_k(\x + \delta) \neq y \\
& & & & & & & D(\x + \delta, \x) \leq \epsilon; \,  \x + \delta \in \X
\end{aligned}
\end{equation}
Alternatively, the objective is to find adversarial examples that are minimally distorted w.r.t a distance function $D$: 
\begin{equation}
\begin{aligned}
\label{eq:minimal_adv_optim}
& \underset{\delta}{\text{min}} & & D(\x + \delta, \x) & & \text{s.t.} & &  \argmax_k f_k(\x + \delta) \neq y \\
& & & & & & &
\x + \delta \in \X
\end{aligned}
\end{equation}

In this work, we are interested in solving \autoref{eq:minimal_adv_optim}, which 
is equivalent to solving \autoref{eq:budget_adv_optim} for every $\epsilon$. Thus, it is a more general but more difficult problem.

\subsection{Equivalent problem}
In the general case, constraint $\x+\delta\in\X$ is not necessarily trivial. However, in our context, this corresponds to a box constraint: $\bm{0}\leq \x+\delta\leq\bm{1}$. We therefore handle it with a simple projection $\mathcal{P}_{[0,1]}$.
For brevity, we will omit this constraint in the rest of the paper.
In the above formulations, $\argmax$ is not differentiable, and therefore not readily amenable to gradient-based optimization. We replace the $\argmax$ constraint with an inequality constraint on the logits,
as done in several works, most notably \cite{carlini2017towards}:
\begin{equation}
\begin{aligned}
\label{eq:minimal_adv_optim_dl}
\underset{\delta}{\text{min}}
&& D(\x + \delta, \x)
&& \text{s.t.}
&&  f_y(\x + \delta) - \max_{k\neq y} f_k(\x + \delta) < 0
\end{aligned}
\end{equation}
While more suited to gradients, this constraint is not scale invariant, as noted in \cite{croce2020reliable}:
extreme scaling of the logits may result in gradient masking. 
We use a slightly modified Difference of Logits Ratio (DLR) for this constraint \cite{croce2020reliable} :
\newcommand{\dlr}{\mathrm{DLR^+}(\z,y)=\frac{\z_y - \max\limits_{i\neq y}\z_i}{\z_{\pi_1}-\z_{\pi_3}}}
\begin{equation}\label{eq:dlr}
    \dlr
\end{equation}
where $\z = f(\x)$ and $\pi$ is the ordering of the elements of $\z$ in decreasing order. This loss is negative if and only if $\x$ is not classified as $y$, and its maximum is 1.
Therefore, we solve the following optimization problem:
\begin{equation}
\begin{aligned}
\label{eq:minimal_adv_optim_dlr}
\underset{\delta}{\text{min}}
&& D(\x + \delta, \x) 
&& \text{s.t.}
&& \mathrm{DLR^+}(f(\x+\delta), y) < 0
\end{aligned}
\end{equation}

\subsection{Distances}

Most attacks in the literature measure the size of the perturbations in terms of $\ell_p$-norms. In this work, we propose an attack that can find minimally perturbed adversarial examples \wrt several distances. We limit this work to four common measures: the $\ell_1$ and $\ell_2$ norms, the CIEDE2000 \cite{sharma2005ciede2000} and the LPIPS distance \cite{zhang2018unreasonable}. The CIEDE2000 color difference \cite{sharma2005ciede2000} is a metric designed to assess the perceptual difference between two colors. It is widely used to evaluate color accuracy of displays and printed materials. This metric was designed to be aligned with the perception of the human eye. Generally, a value smaller than 1 means that the color difference is imperceptible, between 1 and 2 is perceptible through close examination, between 2 and 10 is perceptible at a glance, and above 10 means that the colors are different. This metric is calculated in the CIELAB color space, so a conversion is needed (see \autoref{appendix:ciede2000}). The CIEDE2000 is defined between two color pixels, so we use the image level accumulated version as in \cite{zhao2020towards}. The LPIPS distance \cite{zhang2018unreasonable} is a recently proposed perceptual metric based on the distance between deep features of two images for a chosen model. Zhang \etal showed that this distance aligns with human perception better than other perceptual similarity metrics such as SSIM. We use the LPIPS with AlexNet as in \cite{laidlaw2021perceptual}.

\section{Methodology}

\subsection{General Augmented Lagrangian algorithm}

To describe a minimization problem with one inequality constraint, we use the following notation:
\begin{equation}
\label{eq:general_inequality_problem}
\text{minimize}\quad g(\x) \quad \text{subject to}\quad h(\x) < 0
\end{equation}
with $g:\R^d\rightarrow\R$ the \emph{objective function} and $h:\R^d\rightarrow\R$ the \emph{inequality-constraint function}.

Penalty methods trade a constrained problem \eqref{eq:general_inequality_problem} with an unconstrained one by adding a term (penalty), which increases when the constraint is violated.
A weight of the penalty, $\lambda$, is chosen heuristically, and the unconstrained optimization is repeated several times with increasing values of $\lambda$, until the constraint is satisfied.
Augmented Lagrangian methods have well-established advantages over penalty methods in the general context of optimization \cite{bertsekas2014constrained, bertsekas2016nonlinear, fletcher2013practical, nocedal2006numerical}, and avoid completely such heuristics. They estimate automatically the multipliers, which yields adaptive and optimal weights for the constraints. Such estimates tend to the Lagrange multiplier, which avoids the ill-conditioning problems often encountered in penalty methods \cite{bertsekas2014constrained, conn1997globally}. Augmented Lagrangian methods also avoid the need to explicitly solve the dual problem, unlike basic Lagrangian optimization, which may be intractable/unstable for non-convex problems. From a more practical standpoint, the efficiency of Augmented Lagrangian methods depends solely on the ability to solve the inner minimization (which we detail below), making the implementation simpler and more robust. Despite these advantages and their popularity in the optimization community, Augmented Lagrangian methods have not (to our knowledge) been investigated for adversarial attacks. In the following, we customize Augmented Lagrangian principles to solve adversarial-attack problems of the form \eqref{eq:minimal_adv_optim}. 

\begin{figure}
    \centering
    \includegraphics[width=\columnwidth]{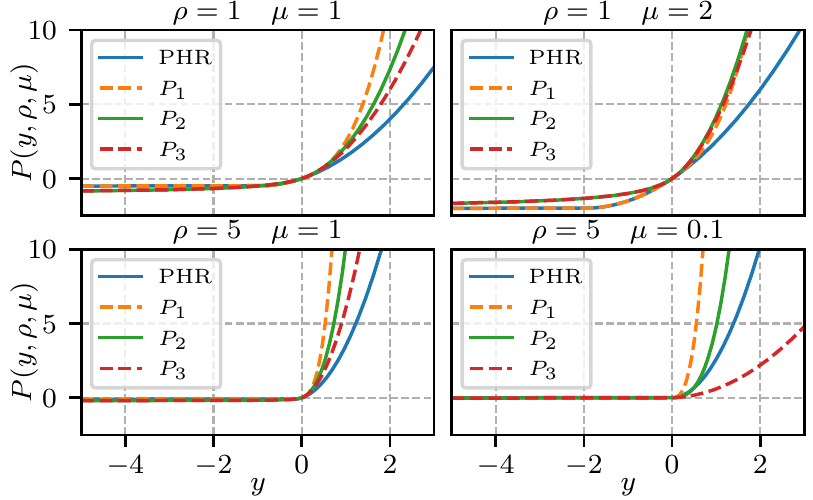}
    \caption{Examples of penalty-Lagrangian functions for different values of $\rho$ and $\mu$. The plotted functions are defined in \cite{birgin2005numerical} and given in \autoref{appendix:penalties}.}
    \label{fig:penalties}
\end{figure}
In general, an Augmented Lagrangian algorithm uses a succession of unconstrained optimization problems, each solved approximately. The algorithm can be broken down in two types of iterations: \emph{outer} iterations indexed by $i$ and \emph{inner} iterations. During the inner iterations, the following Augmented Lagrangian function:
\begin{equation}
    \label{eq:inner_min}
    G(\x) = g(\x) + P(h(\x), \rho^{(i)}, \mu^{(i)})
\end{equation}
is approximately minimized \wrt to $\x$, using the previous solution as initialization, up to an inner convergence criterion. $P:\R\times\R_+^*\times\R_+^*\rightarrow\R$ is a \emph{penalty-Lagrangian function} such that $P'(y,\rho,\mu)=\frac{\partial}{\partial y}P(y,\rho,\mu)$ exists and is continuous for all $y\in\R$ and $(\rho,\mu)\in(\R_+^*)^2$. Any candidate function $P$ should satisfy these four axioms \cite{birgin2005numerical}:
\begin{itemize}[leftmargin=*,label={}]
    \item {\bf Axiom 1.} $\forall y\in\R, \forall (\rho, \mu) \in(\R_+^*)^2, \; \frac{\partial}{\partial y}P(y, \rho, \mu)\geq 0$
    \item {\bf Axiom 2.} $\forall (\rho, \mu) \in(\R_+^*)^2, \; \frac{\partial}{\partial y}P(0, \rho, \mu)=\mu$
    \item {\bf Axiom 3.} If, for all $j\in\N$, $0 < \mu_\text{min} \leq \mu^{(j)} \leq\mu_\text{max} < \infty$, then: $\lim\limits_{j\rightarrow\infty}\rho^{(j)}=\infty$ and $\lim\limits_{j\rightarrow\infty}y^{(j)}=y>0$ imply that $\lim\limits_{j\rightarrow\infty}\frac{\partial}{\partial y}P(y^{(j)}, \rho^{(j)}, \mu^{(j)})=\infty$
    \item {\bf Axiom 4.} If, for all $j\in\N$, $0 < \mu_\text{min} \leq \mu^{(j)} \leq\mu_\text{max} < \infty$, then: $\lim\limits_{j\rightarrow\infty}\rho^{(j)}=\infty$ and $\lim\limits_{j\rightarrow\infty}y^{(j)}=y<0$ imply that $\lim\limits_{j\rightarrow\infty}\frac{\partial}{\partial y}P(y^{(j)}, \rho^{(j)}, \mu^{(j)})=0$
\end{itemize} 
The first and second axioms guarantee that the derivative of the penalty-Lagrangian function $P$ is positive and equal to $\mu$ when $y=0$. The third and fourth axioms guarantee that the derivative of $P$ \wrt $y$ tends to infinity if the constraint is not satisfied (\ie $h(\x)\geq 0$), and tends to 0 if the constraint is satisfied (\ie $h(\x) < 0$). \autoref{fig:penalties} contains examples of widely used penalty-Lagrangian functions. Once the inner convergence criterion is satisfied, an outer iteration is performed, during which the \emph{penalty multiplier} $\mu$ and the \emph{penalty parameter} $\rho$ are modified. The penalty multiplier $\mu$ is updated to 
the derivative of $P$ at the current value \wrt the constraint function:
\begin{equation}
    \label{eq:mu_update}
    \mu^{(i+1)} = P'(h(\x), \rho^{(i)}, \mu^{(i)})
\end{equation}
This means that the penalty multiplier increases when the constraint is not satisfied and, otherwise, is reduced. This can be seen as an {\em adaptive} way of choosing the penalty weight in a penalty method. The penalty parameter $\rho$ is increased based on the value of the constraint function at the current outer iteration, compared to the previous one. If the constraint function has not improved (\ie reduced) significantly, then $\rho$ is multiplied by a fixed factor, typically between 2 and 100 \cite{birgin2005numerical}. With higher values of the penalty parameter $\rho$, the penalty-Lagrangian function tends to an ideal penalty, as can be seen in \autoref{fig:penalties}. \autoref{alg:general_alm} describes a generic Augmented Lagrangian method.

\begin{algorithm}
    \caption{Generic Augmented Lagrangian method}
    \small
    \label{alg:general_alm}
    \begin{algorithmic}[1] 
        \Require Function to minimize $f$, constraint function $g$
        \Require Initial value $\x^{(0)}$
        \Require Penalty function $P$, initial multiplier $\mu^{(0)}$,  $\rho^{(0)}$ 
        \For{$i \gets 0$ to $N-1$}
            \State Using $\x^{(i)}$ as initialization, minimize (approximately): $G(\x) = g(\x) + P(h(\x),\rho^{(i)},\mu^{(i)})$
            \State $\x^{(i+1)} \gets$ approximate minimizer of $G$
            \State $\mu^{(i+1)} \gets P'(h(\x^{(i+1)}), \rho^{(i)}, \mu^{(i)})$
            \If{the constraint does not improve} 
                \State Set $\rho^{(i+1)} > \rho^{(i)}$
            \Else
                \State $\rho^{(i+1)}\gets \rho^{(i)}$
            \EndIf
        \EndFor
    \end{algorithmic}
\end{algorithm}

\subsection{Augmented Lagrangian Attack}

\begin{algorithm}
    \small
    \caption{ALMA attack}
    \label{alg:alm_attack}
    \begin{algorithmic}[1] 
        \Require Classifier $f$, original image $\x$, true or target label $y$
        \Require Number of iterations $N$, initial step size $\eta^{(0)}$, penalty parameter increase rate $\gamma > 1$, constraint improvement rate $\tau\in[0,1]$, $M$ number of steps between $\rho$ increase.
        \Require $D$ distance function
        \Require Penalty function $P$, initial multiplier $\mu^{(0)}$, initial penalty parameter $\rho^{(0)}$ 
        \State Initialize $\xadv^{(0)} \gets \x$, 
        \For{$i \gets 0 $ to $N - 1$}
            \State $\z \gets f(\xadv^{(i)})$
            \State $d^{(i)} \gets \mathrm{DLR^+}(\z,y)$ \textcolor{red}{\Comment{ $\mathrm{tDLR^+}$ for targeted attack}}
            \State $\hat{\mu} \gets \nabla_d P(d^{(i)},\rho^{(i)},\mu^{(i)})$ \textcolor{blue}{\Comment{New penalty multiplier}}
            \State $\mu^{(i+1)} \gets \mathcal{P}_{[\mu_\text{min},\mu_\text{max}]}[\alpha\mu^{(i)}+(1-\alpha)\hat{\mu}]$ \textcolor{blue}{\Comment{EMA}}
            \State $L \gets D(\xadv^{(i)}, \x) + P(d^{(i)}, \rho^{(i)}, \mu^{(i+1)})$ \textcolor{blue}{\Comment{Loss}}
            \State $\bm{g} \gets \nabla_{\xadv}L$ \textcolor{blue}{\Comment{Gradient of loss \wrt $\xadv$}}
            \State $\xadv^{(i+1)} \gets \mathcal{P}_{[0,1]}[\xadv^{(i)} - \eta^{(i)} \bm{g}]$ \textcolor{blue}{\Comment{Step and box-constraint}}
            \If{$(i+1)\bmod M =0$\; and \;$d^{(j)} > 0, \forall j\in\{0,\dots,i\}$\\\quad\, \textbf{and} \;$d^{(i)}>\tau d^{(i-M)}$}
                \State $\rho^{(i+1)} \gets \gamma \rho^{(i)}$ \textcolor{blue}{\Comment{\parbox[t][0em]{.4\linewidth}{If no adversarial has been found and $d$ does not decrease significantly, increase $\rho$ by a factor of $\gamma$}}}
            \Else   
                \State $\rho^{(i+1)} \gets \rho^{(i)}$
            \EndIf
        \EndFor
        \State Return $\xadv^{(i)}$ that is adversarial and has smallest $D(\xadv^{(i)}, \x)$
    \end{algorithmic}
\end{algorithm}

We propose to use Augmented Lagrangian methods to solve \eqref{eq:minimal_adv_optim_dlr}. However, we found that a vanilla Augmented Lagrangian algorithm is not well-suited for adversarial attacks. Indeed, alternating between inner and outer iterations is rather slow compared to the few hundreds iterations in typical adversarial attacks. Moreover, we wish to obtain an algorithm with a fixed number of iterations for practical purposes. This is clearly incompatible with the use of an inner convergence criterion required to stop the approximate minimization of $G$ (step 2 of \autoref{alg:general_alm}). Designing a good inner convergence criterion is not a trivial task either. Therefore, we propose a modification of the traditional Augmented Lagrangian algorithm. We combine the inner and outer iterations, resulting in a joint update of the perturbation $\delta=\tilde{\x} - \x$ and the penalty multiplier $\mu$. This means that we do not need an inner convergence criterion as we are always adapting $\mu$. This also requires to adapt the increase of $\rho$, which depends on the value of the inequality-constraint function. \autoref{alg:alm_attack} presents our approach and, in the following sections, we detail several important design choices for the algorithm.

\subsubsection{Penalty parameters adaptation}
The most critical design choices for our attack are the adaptation of the penalty parameters $\mu$ and $\rho$. 

\begin{figure}
    \centering
    \includegraphics[width=\columnwidth]{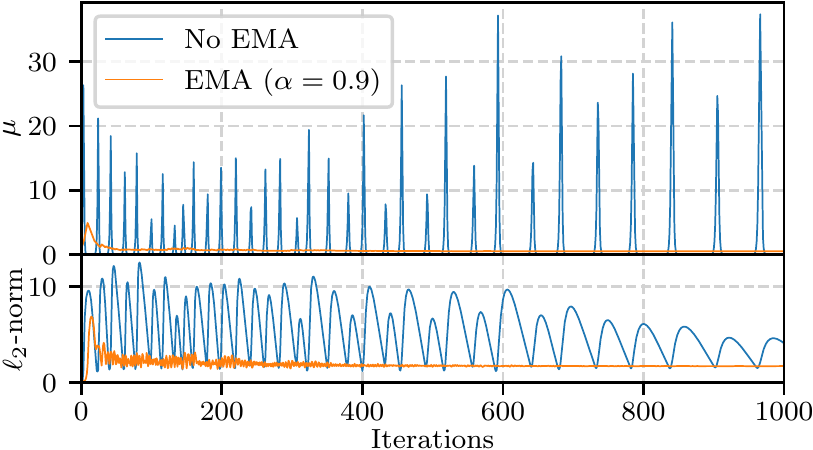}
    \caption{Example of the evolution of the penalty multiplier $\mu$ and the $\ell_2$-norm of the perturbation during the optimization when attacking a single MNIST sample for the SmallCNN model. This leads to a $\ell_2$-norm of the final adversarial perturbation of 2.13 without EMA and 1.71 with EMA ($\alpha=0.9$). The norm of the perturbation can be lower without EMA in some iterations, but it is associated with an increase in $\mu$, meaning that the perturbed sample $\xadv^{(i)}$ is not adversarial.}
    \label{fig:ema_plot}
\end{figure}

\noindent \textbf{Penalty multiplier.}\quad $\mu$ is usually modified in each outer iteration (after the inner problem is approximately solved), and taken as the derivative of the penalty function \wrt the constraint function (see \autoref{eq:mu_update}). In our algorithm, we modify $\mu$ at each iteration. This approach aims at reducing the budget needed for the optimization by combining inner and outer iterations. However, this leads to spiking values of $\mu$, which tend to make the optimization unstable.
\autoref{fig:ema_plot} shows the evolution of $\mu$ during the attack, as well as the $\ell_2$-norm of the perturbation, when attacking (with ALMA $\ell_2$) a single MNIST sample with the SmallCNN model. We can see that $\mu$ regularly spike to high values when not using EMA, which in turns increases the $\ell_2$-norm of the perturbation because the penalty term dominates.
To solve this issue, we smooth the values of $\mu$ using an Exponential Moving Average (EMA) of $\mu$:
\begin{equation}
\begin{aligned}
    \hat{\mu} &= P'(g(\x), \rho^{(i)}, \mu^{(i)})\\
    \mu^{(i+1)} &= \alpha \mu^{(i)} + (1 - \alpha) \hat{\mu}
\end{aligned}
\end{equation}
where $\alpha\in\R$ is a hyper-parameter. This corresponds to steps 5 and 6 in \autoref{alg:alm_attack}. 
When using an EMA with $\alpha=0.9$, the evolution of $\mu$ is smoother, with a much smaller spike at the beginning and then stabilizing. As a consequence, the $\ell_2$-norm of the final perturbation is smaller when using an EMA at 1.71 for $\alpha=0.9$, compared to 2.13 when $\alpha=0$.

For numerical stability, we also need to constrain the value of the penalty multiplier such that it never becomes too small or too large. Therefore, we also project $\mu$ on the safeguarding interval $[\mu_\text{min}, \mu_\text{max}]\subset\R_+^*$. 

\noindent \textbf{Penalty parameter.}\quad $\rho$ is typically increased in each outer iteration if the constraint has not improved during the inner minimization. Since inner and outer iterations are combined in our attack, we adopt a simple strategy to adapt $\rho$. Every $M$ iterations, we verify if an adversarial example was found and, if not, if the constraint has improved. If no adversarial example was found, and the constraint does not improve (\eg reduced by $5\%$ in the last $M$ iterations), we set $\rho^{(i+1)} = \gamma \rho^{(i)}$. This corresponds to steps 10 to 14 of \autoref{alg:alm_attack}.

\subsubsection{Choice of penalty function}
Experiments have shown that the choice of the penalty function is of great importance, especially when considering non-convex problems \cite{birgin2005numerical}. Many functions have been proposed in the literature. In our experiment, we use the $P_2$ penalty function proposed in \cite{kort1976combined}, which is defined as followed:
\begin{equation}
    P_2(y, \rho, \mu) = 
    \begin{cases}
        \mu y + \mu \rho y^2 + \frac{1}{6}\rho^2 y^3 &\text{if}\quad y \geq 0\\
        \frac{\mu y}{1 - \rho y}&\text{if}\quad y < 0\\
    \end{cases}
\end{equation}
where $y$ represents the value of the constraint: $\mathrm{DLR^+}(\z,y)$ in our case (or $\mathrm{tDLR^+}(\z,y)$ in a targeted scenario). In the numerical comparison of several penalty-Lagrangian functions done by Birgin \etal \cite{birgin2005numerical}, $P_2$ is second to PHRQuad in terms of robustness. However, we found that $P_2$ is more suited to our problem. Experimentally, PHRQuad's derivative is smaller than $P_2$, resulting in smaller increase of $\mu$ in each iteration. This, in turn, results in an increase of $\rho$ because no adversarial example is found (see step 10 and 11 of \autoref{alg:alm_attack}). Generally, increasing $\rho$ helps in finding a feasible point (\ie adversarial examples), at the cost of a larger final distance for the perturbation. The choice of $P_2$ also depends on the algorithm design. In our attack, we chose to increase $\rho$ every $M$ steps if no adversarial example has been found, regardless of the usual convergence criterion used in more traditional Augmented Lagrangian methods. Therefore, we need a penalty function with an ``aggressive" derivative such that $\mu$ is modified quickly.

\subsubsection{Learning rate scheduling}
\label{sec:lr_scheduling}

\noindent \textbf{Initial learning rate.}\quad Different distance functions can have widely different scales (see \Autoref{tab:mnist_results,tab:cifar10_imagenet_results}, median results column). Therefore, the initial learning rate is chosen adaptively for each sample $\x$, such that the first gradient step (step 9 in \autoref{alg:alm_attack}) increases $D$ by $\epsilon$, or formally, such that: $D(\tilde{\x}^{(1)}, \x) = \epsilon$. To obtain that value, we compute the gradient $\g$ of $\mathrm{DLR^+}$ \wrt $\x$, and find the scalar $\eta^{(0)}$ such that $D(\mathcal{P}_{[0,1]}[\x-\eta^{(0)}\g],\x)=\epsilon$, using a line search followed by a binary search.

\begin{figure}
    \centering
    \includegraphics[width=\columnwidth]{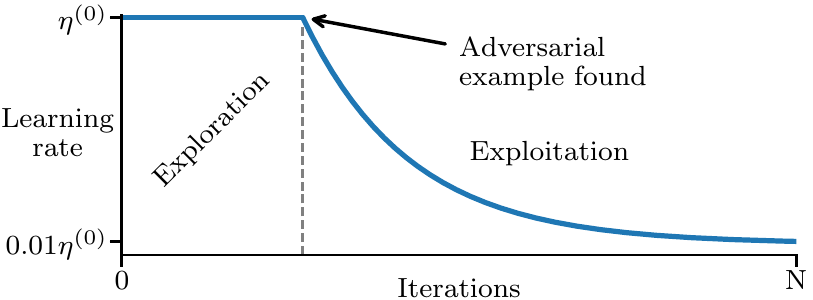}
    \caption{Exponential learning rate decay for the attack.}
    \label{fig:exp_decay}
\end{figure}
\noindent \textbf{Learning rate decay.}\quad The optimization process can be partitioned in two phases: exploration and exploitation. The exploration phase corresponds to finding a feasible point (\ie an adversarial example) for \autoref{eq:minimal_adv_optim}. Then, the exploitation phase consists in refining the feasible point, \ie finding a minimally distorted adversarial example. In our algorithm, the learning rate remains constant during the exploration phase, and decays exponentially during the exploitation phase to reach a final learning rate of $\nicefrac{\eta^{(0)}}{100}$. \autoref{fig:exp_decay} illustrates the scheduling of the learning rate decay.

\subsubsection{Optimization algorithm}
It is well known that gradient descent has a slow convergence rate, especially for ill-conditioned problem. Several optimization algorithms have been proposed to accelerate first-order methods, such as the momentum method or Nesterov's acceleration \cite{nesterov1983method}. One such algorithm is RMSProp \cite{tieleman2012lecture}, in which the learning rate is divided by the square root of the EMA of the squared gradient. We use RMSProp combined with the momentum method in our attack, with a slight modification. Originally, the EMA of the squared gradient is initialized at 0. To avoid dividing by a small value in the first iteration (which would result in a large increase of the distance during the first iteration), we initialize the EMA of the squared gradient at 1. For simplicity, \autoref{alg:alm_attack} only describes a regular gradient descent update at step 9.

\section{Experiments}
\label{sec:experiments}

\noindent \textbf{Datasets.}\quad To evaluate our attack, we conduct experiments on three datasets: MNIST, CIFAR10 and ImageNet with two different budgets: 100 and 1000 iterations. On MNIST, we test the attacks on the whole test set. On CIFAR10 and ImageNet, we test the attacks on 1000 samples. These 1000 samples correspond to the first 1000 samples of the test set on CIFAR10 and 1000 randomly chosen samples from the validation set for ImageNet.


\noindent \textbf{Models.}\quad It has been observed that some attacks can work well for regularly trained models, and fail against defended (\ie adversarially trained) models \cite{carlini2019evaluating, croce2020reliable, rony2019decoupling}. Therefore, for each dataset, we evaluate the attack against a collection of models. Specifically, for MNIST, we use four models. The first three are the same model as in \cite{carlini2017towards, pintor2021fast, rony2019decoupling} with three training schemes: a regularly trained model we call SmallCNN, a $\ell_2$-adversarially trained model from \cite{rony2019decoupling} we call SmallCNN-DDN and a $\ell_{\infty}$-adversarially trained model from \cite{zhang2019theoretically} we call SmallCNN-TRADES. To vary architectures, we also test on the $\ell_\infty$-adversarially trained model from \cite{zhang2019towards} we call CROWN-IBP. This is the large variant trained with $\epsilon=0.4$.
For CIFAR10, we use three models: a regularly trained Wide ResNet 28-10 \cite{zagoruyko2016wide}, a $\ell_{\infty}$-adversarially trained Wide ResNet 28-10 from \cite{carmon2019unlabeled} and a $\ell_2$-adversarially trained ResNet-50 \cite{he2016deep} from \cite{augustin2020adversarial}.
These models are fetched from the RobustBench library \cite{croce2020robustbench}.
On ImageNet, we test the attacks on three ResNet-50 models: one regularly trained, and two adversarially trained ($\ell_2$ and $\ell_{\infty}$), available through the robustness library \cite{robustness}. All the models have been selected because they have obtained high robust accuracies in third-party evaluations \cite{croce2020reliable}.

\noindent \textbf{Metrics.}\quad For each attack, we report the Attack Success Rate (ASR) which is the proportion of examples for which an adversarial perturbation was found, and the median perturbation size. To compare the complexities of the attacks, we use the average number of forward and backward model propagations per sample needed to perform each attack. In the deep learning context, the number of model propagations (forward and backward) is a good proxy for the complexity of an attack, as these operations generally require orders of magnitude more computations than the other operations of an attack. We also prefer this measure to the attack run-time since this makes for comparisons that are independent of both software optimizations and hardware differences.

\noindent \textbf{Hyperparameters.}\quad For our attack, we consider two budgets: 100 and 1000 iterations. The goal of this work is to propose a framework to generate minimally perturbed adversarial example \wrt any distance. Therefore, we chose one $\alpha$ (\ie the coefficient of the EMA) per budget and one $\epsilon$ (\ie the increase of $D$ in the first iteration, see \autoref{sec:lr_scheduling}) per distance. $\alpha$ and $\epsilon$ are shared accross all datasets and models. For the 100 iterations budget, we set $\alpha=0.5$ and for the 1000 iterations budget, we set $\alpha=0.9$. For $\epsilon$, we chose an initial value in relation to the usual scale of each distance: typically a tenth of the expected distances between the adversarial examples and the original inputs works well. Although it requires prior knowledge on the expected distances, a good $\epsilon$ can quickly be found when experimenting with a new distance function. \autoref{appendix:epsilon} gives the initial values of $\epsilon$ used in our experiments for each distance. The other hyperparameters are fixed. We use $\mu^{(0)}=\rho^{(0)}=1$, $\mu_\text{min}=10^{-6}$ and $\mu_\text{max}=10^{12}$, $\gamma=1.2$, $\tau=0.95$ and $M=10$ in all our experiments. These values are not usual for Augmented Lagrangian algorithms; they reflect our design choice of combining inner and outer iterations. Since we update our penalty parameters (\ie $\mu$ and $\rho$) more frequently, $\gamma$ can have a smaller value.

\noindent \textbf{Attacks.}\quad We compare our attack with various state-of-the-art attacks from the literature. Specifically, we evaluate the $\ell_1$ variant of our attack against the EAD attack \cite{chen2018ead} with budgets of 9$\times$100 and 9$\times$1000, and FAB $\ell_1$ \cite{croce2020minimally} and FMN $\ell_1$ \cite{pintor2021fast} attacks with budgets of 100 and 1000 iterations. For the $\ell_2$ variant of our attack, we compare it to the C\&W $\ell_2$ \cite{carlini2017towards} attack with budgets of 9$\times$1000 and 9$\times$10\,000, DDN \cite{rony2019decoupling}, FAB $\ell_2$ \cite{croce2020minimally} and  FMN $\ell_2$ \cite{pintor2021fast} all with budgets of 100 and 1000 iterations.\footnote{We tried to include the B\&B $\ell_1$ and $\ell_2$ attacks \cite{brendel2019accurate} in our comparison, but their official implementations kept crashing. In the cases where it did not, we observed median distances worse than FAB. As such, we omitted the method completely from the experiments.} For FAB $\ell_1$ and $\ell_2$ on ImageNet, we use the targeted variant of the attack, as was done in the original work.\footnote{The FAB attack needs to compute the Jacobian of the output of the model \wrt to the input. This increases the complexity by a factor K (\ie the number of classes), making it impractical for datasets with large number of classes, such as ImageNet. See Section 4 of \cite{croce2020minimally}.}  We compare the CIEDE2000 variant of our attack with, to the best of our knowledge, the only attack using this metric: the PerC-AL attack \cite{zhao2020towards} with budgets of 100 and 1000 iterations. Lastly, for the LPIPS variant, we compare our attack to the LPA attack \cite{laidlaw2021perceptual}, which is designed to solve \autoref{eq:budget_adv_optim} (\ie find a perturbation within a distance budget). We perform a binary search on the distance budget to find the minimum budget for which an adversarial example can be found. We also add the APGD$^\text{T}_\text{DLR}$ $\ell_2$ \cite{croce2020reliable} to the $\ell_2$ comparison with a binary search. For the binary seaarch, we start with a large enough budget (depending on the dataset) such that the attacks can reach 100\% ASR and perform enough binary search steps to reach a precision of 0.01 for the $\ell_2$-norm and 0.001 for the LPIPS.

Finally, we perform a C\&W type attack as a baseline for the CIEDE2000 and LPIPS distances. We replace the $\ell_2$-norm with the corresponding target distance and use a budget of 9$\times$1000 which is enough to get a good performance while keeping an acceptable budget (compared to 9$\times$10\,000).

\section{Results}

\begin{table}
    \centering
    \resizebox{\columnwidth}{!}{
    \begin{tabular}{clccc}
    Distance & Attack & ASR (\%) & \makecell{Median\\Distance} & \makecell{Forwards /\\Backwards} \\
    \midrule[1pt]
    \multirow{8}{*}{$\ell_1$-norm}
        & EAD 9$\times$100 \cite{chen2018ead} & 97.18 & 21.94 & 807 / 407 \\ 
        & EAD 9$\times$1000 \cite{chen2018ead} & 97.76 & 19.19 & 5\,025 / 2\,516 \\
        & FAB $\ell_1$ 100 \cite{croce2020minimally} & 99.80 & 27.41 & 201 / 1\,000 \\
        & FAB $\ell_1$ 1000 \cite{croce2020minimally} & 99.83 & 24.21 & 2\,001 / 10\,000 \\
        & FMN $\ell_1$ 100 \cite{pintor2021fast} & 69.87 & -- & 100 / 100 \\
        & FMN $\ell_1$ 1000 \cite{pintor2021fast} & 95.35 & 7.34 & 1\,000 / 1\,000 \\
        & ALMA $\ell_1$ 100 & 99.90 & 11.45 & 100 / 100\\
        & ALMA $\ell_1$ 1000 & 100 & 7.16 & 1\,000 / 1\,000\\
    \midrule[0.75pt]
    \multirow{11}{*}{$\ell_2$-norm}
        & C\&W $\ell_2$ 9$\times$1000 \cite{carlini2017towards} & 40.19 & -- & 8\,643 / 8\,643 \\
        & C\&W $\ell_2$ 9$\times$10\,000 \cite{carlini2017towards} & 40.20 & -- & 85\,907 / 85\,907 \\
        & DDN 100 \cite{rony2019decoupling} & 98.48 & 1.86 & 100 / 100 \\
        & DDN 1000 \cite{rony2019decoupling} & 99.83 & 1.61 & 1\,000 / 1\,000 \\
        & FAB $\ell_2$ 100 \cite{croce2020minimally} & 83.25 & 2.10 & 201 / 1\,000 \\
        & FAB $\ell_2$ 1000 \cite{croce2020minimally} & 96.28 & 1.77 & 2\,001 / 10\,000 \\
        & FMN $\ell_2$ 100 \cite{pintor2021fast} & 70.99 & 3.18 & 100 / 100 \\
        & FMN $\ell_2$ 1000 \cite{pintor2021fast} & 96.02 & 1.77 & 1\,000 / 1\,000 \\
        & APGD$^\text{T}_\text{DLR}$ $\ell_2$\textsuperscript{\ddag} \cite{croce2020reliable} & 99.98 & 2.52 & 12\,271 / 12\,253\\
        & ALMA $\ell_2$ 100 & 99.72 & 2.38 & 100 / 100\\
        & ALMA $\ell_2$ 1000 & 100 & 1.61 & 1\,000 / 1\,000\\
    \bottomrule[1pt]
    \end{tabular}}
    \caption{Performance for attacks on the MNIST dataset. Geometric mean over the four models. For C\&W $\ell_2$, the attack fails on the IBP model (3\% ASR), so the median distance is undefined. \textsuperscript{\ddag}A binary search is performed on each sample to get a minimal perturbation attack (\autoref{eq:minimal_adv_optim}).}
    \label{tab:mnist_results}
\end{table}

\begin{table*}
    \centering
    \footnotesize
    \begin{tabular}{clccccccc}
     & & \multicolumn{3}{c}{CIFAR10} & & \multicolumn{3}{c}{ImageNet} \\
    \cmidrule[0.75pt]{3-5}\cmidrule[0.75pt]{7-9}
    Distance & Attack & ASR (\%) & \makecell{Median\\Distance} & \makecell{Forwards /\\Backwards} & & ASR (\%) &  \makecell{Median\\Distance} & \makecell{Forwards /\\Backwards} \\
    \midrule[1pt] 
    \multirow{8}{*}{$\ell_1$-norm}
        & EAD 9$\times$100 \cite{chen2018ead} (AAAI'17) & 100 & 6.11 & 572 / 290 & & 100 & 13.87 & 488 / 248\\
        & EAD 9$\times$1000 \cite{chen2018ead} (AAAI'17) & 100 & 5.44 & 4\,284 / 2\,146 & & 100 & 12.83 & 3\,758 / 1\,883\\
        & FAB\textsuperscript{\dag} $\ell_1$ 100 \cite{croce2020minimally} (ICML'20) & 96.58 & 4.26 & 201 / 1\,000 & & 88.82 & 10.72 & 1\,810 / 900\\
        & FAB\textsuperscript{\dag} $\ell_1$ 1000 \cite{croce2020minimally} (ICML'20) & 99.00 & 3.78 & 2\,001 / 10\,000 & & 89.07 & 8.88 & 18\,010 / 9\,000 \\
        & FMN $\ell_1$ 100 \cite{pintor2021fast} & 99.90 & 3.64 & 100 / 100 & & 94.33 & 8.43 & 100 / 100 \\
        & FMN $\ell_1$ 1000 \cite{pintor2021fast} & 99.83 & 3.54 & 1\,000 / 1\,000 & & 93.93 & 7.58 & 1\,000 / 1\,000 \\
        & ALMA $\ell_1$ 100 & 100 & 4.31 & 100 / 100 & & 100 & 19.79 & 100 / 100\\
        & ALMA $\ell_1$ 1000 & 100 & 3.69 & 1\,000 / 1\,000 & & 100 & 12.10 & 1\,000 / 1\,000\\
    \midrule[0.75pt]
    \multirow{11}{*}{$\ell_2$-norm}
        & C\&W $\ell_2$ 9$\times$1000 \cite{carlini2017towards} (SP'17) & 100 & 0.40 & 7\,976 / 7\,974 & & 99.83 & 0.57 & 7\,248 / 7\,246 \\
        & C\&W $\ell_2$ 9$\times$10\,000 \cite{carlini2017towards} (SP'17) & 100 & 0.40 & 78\,081 / 78\,079 & & 99.83 & 0.57 & 67\,479 / 67\,476 \\
        & DDN 100 \cite{rony2019decoupling} (CVPR'19) & 100 & 0.43 & 100 / 100 & & 99.70 & 0.51 & 100 / 100\\
        & DDN 1000 \cite{rony2019decoupling} (CVPR'19) & 100 & 0.42 & 1\,000 / 1\,000 & & 99.87 & 0.50 & 1\,000 / 1\,000\\
        & FAB\textsuperscript{\dag} $\ell_2$ 100 \cite{croce2020minimally} (ICML'20) & 100 & 0.41 & 201 / 1\,000 & & 99.70 & 0.35 & 1\,810 / 900\\
        & FAB\textsuperscript{\dag} $\ell_2$ 1000 \cite{croce2020minimally} (ICML'20) & 100 & 0.41 & 2\,001 / 10\,000 & & 98.90 & 0.35 & 18\,010 / 9\,000\\
        & FMN $\ell_2$ 100 \cite{pintor2021fast} & 99.90 & 0.43 & 100 / 100 & & 99.43 & 0.38 & 100 / 100 \\
        & FMN $\ell_2$ 1000 \cite{pintor2021fast} & 99.83 & 0.40 & 1\,000 / 1\,000 & & 99.63 & 0.36 & 1\,000 / 1\,000 \\
        & APGD$^\text{T}_\text{DLR}$ $\ell_2$\textsuperscript{\ddag} \cite{croce2020reliable} (ICML'20) & 100 & 0.38 & 5\,345 / 5\,321 & & 100 & 0.34 & 6\,096 / 6\,068\\
        & ALMA $\ell_2$ 100 & 100 & 0.40 & 100 / 100 & & 100 & 0.38 & 100 / 100\\
        & ALMA $\ell_2$ 1000 & 100 & 0.38 & 1\,000 / 1\,000 & & 100 & 0.35 & 1\,000 / 1\,000\\
    \midrule[0.75pt]
    \multirow{5}{*}{\makecell{CIEDE\\2000}}
        & C\&W CIEDE2000 9$\times$1000 & 100 & 0.93 & 6\,729 / 6\,726 & & 100 & 1.39 & 5\,635 / 5\,632 \\
        & PerC-AL 100 \cite{zhao2020towards} (CVPR'20)& 100 & 2.87 & 201 / 100 & & 99.90 & 3.55 & 201 / 100\\
        & PerC-AL 1000 \cite{zhao2020towards} (CVPR'20) & 100 & 2.72 & 2\,001 / 1\,000 & & 99.93 & 3.42 & 2\,001 / 1\,000\\
        & ALMA CIEDE2000 100 & 100 & 1.09 & 100 / 100 & & 100 & 0.75 & 100 / 100\\
        & ALMA CIEDE2000 1000 & 100 & 0.78 & 1\,000 / 1\,000 & & 100 & 0.63 & 1\,000 / 1\,000\\
    \midrule[0.75pt]
    \multirow{4}{*}{\makecell{LPIPS\\$\times 10^-2$}}
        & C\&W LPIPS 9$\times$1000 & 100 & 0.47 & 6\,658 / 6\,655 & & 100 & 2.07 & 4\,950 / 4\,944 \\
        & LPA\textsuperscript{\ddag} \cite{laidlaw2021perceptual} (ICLR'21) & 100 & 5.39 & 1\,118 / 1\,108 & & 100 & 5.79 & 1\,211 / 1\,201 \\
        & ALMA LPIPS 100 & 99.97 & 2.47 & 100 / 100 & & 100 & 1.59 & 100 / 100\\
        & ALMA LPIPS 1000 & 100 & 0.60 & 1\,000 / 1\,000 & & 100 & 1.13 & 1000 / 1000\\
    \bottomrule[1pt]
    \end{tabular}
    \caption{Performance for attacks on the CIFAR10 and ImageNet datasets. Geometric mean over the three models for each dataset. \textsuperscript{\dag}For ImageNet, we use the targeted variant of the attack as in \cite{croce2020minimally} (see \autoref{sec:experiments}). \textsuperscript{\ddag}A binary search is performed on each sample to get a minimal perturbation attack (\autoref{eq:minimal_adv_optim}).}
    \label{tab:cifar10_imagenet_results}
\end{table*}

\begin{figure}
    \centering
    \includegraphics[width=\columnwidth]{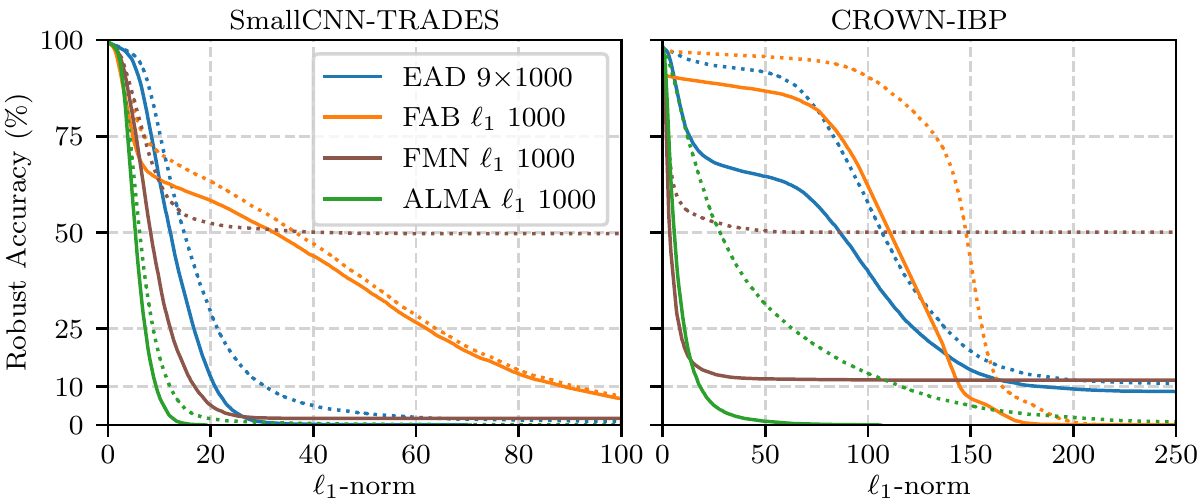}\\
    \includegraphics[width=\columnwidth]{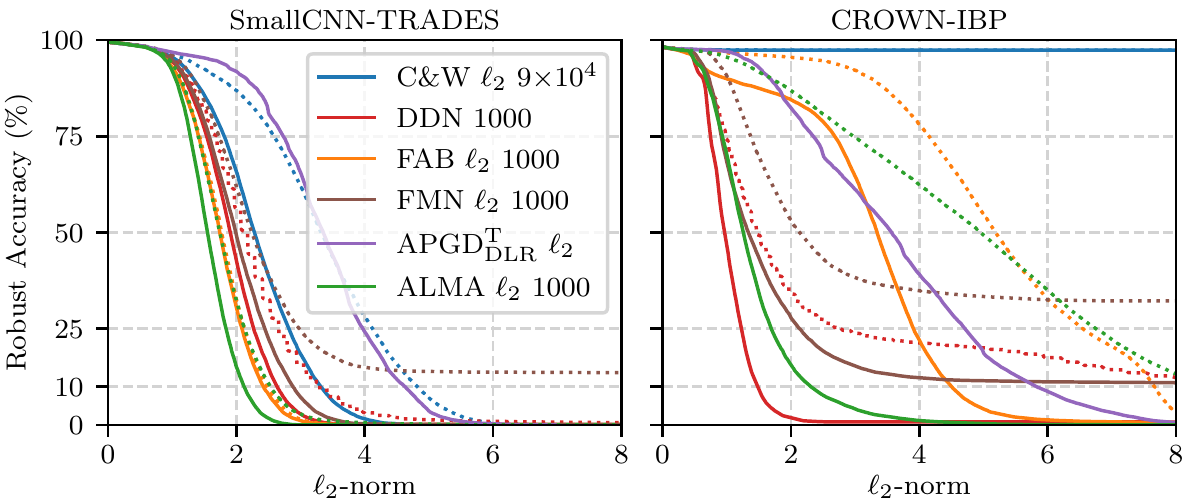}
    \caption{Robust accuracy curves for the SmallCNN-TRADES and CROWN-IBP adversarially trained models on MNIST against $\ell_1$ (top row) and $\ell_2$ (bottom row) attacks.
    }
    \label{fig:mnist_curves_combined}
\end{figure}


\begin{figure}
    \centering
    \includegraphics[width=\columnwidth]{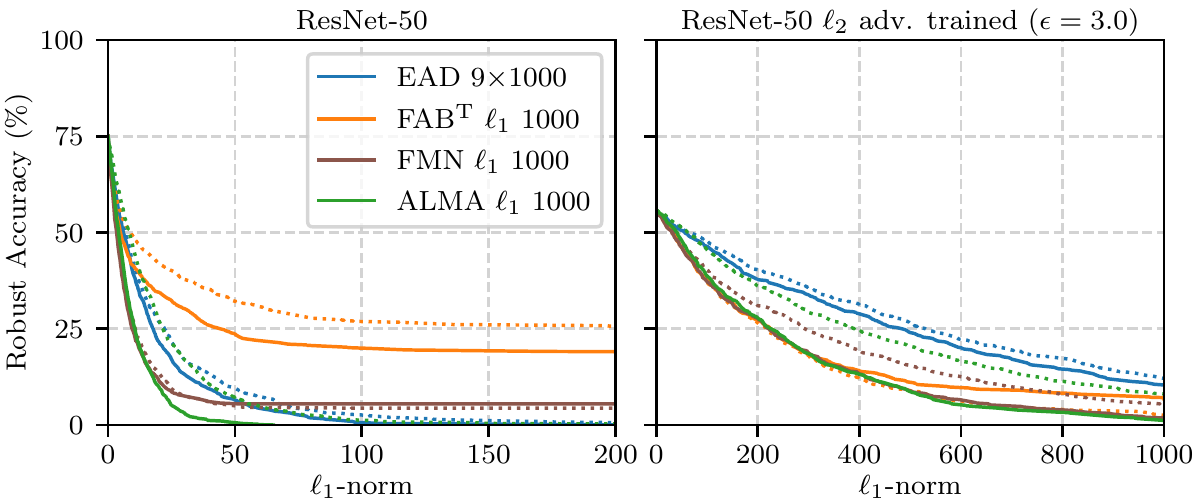}\\
    \caption{Robust accuracy curves for regular and $\ell_2$-adversarial ResNet-50 on ImageNet against $\ell_1$ attacks.
    }
    \label{fig:imagenet_curves_combined}
\end{figure}

To summarize the results, we report the geometric mean of each metric over the models considered for each dataset, in \autoref{tab:mnist_results} for MNIST and \autoref{tab:cifar10_imagenet_results} for CIFAR10 and ImageNet. Tables containing the detailed results for each model can be found in \autoref{appendix:detailed_results}. We also present robust accuracy curves, which reflect directly the performance in terms of minimum distance of the adversarial examples, by showing an expected robust accuracy as a function of a perturbation budget. For all curves, the dotted lines denote the reduced budget attacks of the corresponding colors. With these curves, we can also find the median distance (see \autoref{appendix:detailed_results}) for each attack by looking at the distance for which the accuracy is 50\%. All the robust accuracy curves can be found in \autoref{appendix:curves}. Here we display a few for which there are significant differences between the attacks. For MNIST, we show the curves in \autoref{fig:mnist_curves_combined} for the SmallCNN-TRADES and CROWN-IBP models against $\ell_1$ and $\ell_2$ attacks. For ImageNet, we show the curves for ResNet-50 regularly and $\ell_2$-adversarially trained from \cite{robustness} against $\ell_1$ attacks.

Across all datasets and models, our attack consistently obtains results competitive with attacks tailored to each distance. On MNIST, ALMA $\ell_1$ is the only attack to reach an ASR of 100\% with only FAB $\ell_1$ outperforming it on the SmallCNN model. FMN $\ell_1$ reaches a lower median $\ell_1$ at the cost of a reduced ASR on the CROWN IBP model.
ALMA $\ell_2$ is only marginally outperformed by APGD$^\text{T}_\text{DLR}$ $\ell_2$ on the SmallCNN and SmallCNN-DDN models, but again, ALMA is the only attack to consistently reach 100\% ASR.
The results on MNIST get confirmed by the experiments on CIFAR10 and ImageNet. All the variants of ALMA consistently obtain 100\% success rate (except for ALMA LPIPS 100). ALMA $\ell_1$ obtains worse median norms than FAB $\ell_1$ and FMN $\ell_1$, but again, at the cost of a reduced ASR for both attacks. Surprisingly, FMN $\ell_1$ obtains lower ASR for the higher budget variant, but with a lower median. For the $\ell_2$ attacks, several perform similarly, with only APGD$^\text{T}_\text{DLR}$ $\ell_2$ reaching a lower $\ell_2$ median with a 100\% ASR. This is expected given that this is a distance budget attack, with a much higher overall complexity when combining it with a binary search. 
For the CIEDE2000 distance, the PerC-AL attack \cite{zhao2020towards} obtains significantly larger median distance on both CIFAR10 and ImageNet. Investigating the original code, we found errors in the implementation of the CIEDE2000, resulting in both wrong values and wrong gradients.\footnote{To verify this, we tested both implementations of the CIEDE2000 against the test values provided for this purpose in \cite{sharma2005ciede2000}.}.
The LPIPS variant of ALMA performs much better than LPA combined with a binary search for both datasets.
Being a penalty base approach, our C\&W baseline gets performance that is on par or better than ALMA LPIPS on both datasets, but at a much higher computational cost. However, it does not perform as well for the CIEDE2000 distance.

Overall, our attack offers a reliable trade-off between speed and performance in terms of ASR and perturbation size, given that the hyper-parameters are not tuned beyond $\epsilon$, which is distance specific, and $\alpha$ which is directly related to the number of steps.

\section{Conclusion}
In this paper, an adversarial attack based on Augmented Lagrangian methods is proposed, which acts as a general framework for generating minimally perturbed adversarial examples \wrt several distances. In most of our experiments, it offers a good trade-off between performance and computational complexity in comparison to state-of-the-art methods. We believe that our general method could serve as a starting point for designing efficient attacks minimizing new distances.

{\small
\bibliographystyle{ieee_fullname}
\bibliography{egbib}
}

\clearpage
\appendix
\section*{\begin{Large}Appendix\end{Large}}

\section{Differences with ADMM attack \cite{zhao2019admm} and StrAttack \cite{xu2018structured}}

There is a significant technical difference between our proposed attack and the two attacks based on ADMM approaches \cite{zhao2019admm, xu2018structured} as the multipliers in the ADMM attacks are used for the problem-splitting constraints, but not for the attack constraints as in our ALM. For two terms, ADMM replaces a one-variable problem of the form $\min_x f(x) + g(x)$ by a two-variable problem:
\begin{equation}
    \min_{x,y} f(x) + g(y) \quad \text{s.t.} \quad x=y
\end{equation}
a splitting that gives raise to variable-consistency constraints $x=y$.
In fact, both ADMM attacks \cite{xu2018structured, zhao2019admm} are based on a decomposition of the Carlini-Wagner penalty formulation (Equation 7 in \cite{zhao2019admm} and Equation 4 in \cite{xu2018structured}): 
\begin{equation}
    D(x+\delta, x) + g_{\tiny{\text{CW}}} (\beta) \quad \text{s.t.} \quad \delta = \beta
\end{equation}
where $D$ is the distance and $g_{\tiny{\text{CW}}}$ is the standard CW penalty for attack constraint $f_y(\x + \delta) - \max_{k\neq y} f_k(\x + \delta) < 0$; see Equation 5 in \cite{zhao2019admm} and Equation 3 in \cite{xu2018structured}. The Lagrange multipliers in these ADMM attacks are for the decomposition constraints $\delta = \beta$, but the attack constraints are still handled with the standard CW penalty. In our case, we address the attack constraints with augmented Lagrangian principles, and there is no ADMM splitting in our method. 
The ADMM attack in \cite{zhao2019admm} has no public implementation, so we were not able to implement it in our experimental framework. The publicly available implementation of the StrAttack \cite{xu2018structured} contains several differences with the original paper regarding hyper-parameters and update rules for the auxiliary variables. Therefore, we contacted the authors of both papers (of which several are in common) regarding the lack of public implementation, and discrepancies between paper and code, but did not get any answer. Therefore, we did not include these attacks in our experiments. It should be noted that StrAttack's \cite{xu2018structured} implementation is based on the C\&W $\ell_2$ attack, but adds a sparsity objective, which tends to increase the perturbation size in terms of $\ell_2$ norm compared to the vanilla C\&W $\ell_2$ attack.

\section{CIEDE2000}
\label{appendix:ciede2000}

The CIEDE2000 color difference formula is complex, so we advise the reader to look at the original work \cite{sharma2005ciede2000}. This formula is calculated using the CIELAB color space. However, most image datasets are provided in an RGB format. Therefore, to use the CIEDE2000 color difference formula, we must first convert the images from RGB to the CIELAB color space. We need to use a first conversion between RGB and XYZ color spaces. For this step, we need to know the RGB working space and the reference white. However, we do not have that information available, as we do not know how the images were captured in the first place. As a consequence, we make the assumption of the sRGB working space with an Illuminant D65 white reference. With these assumptions, the formula to convert from RGB (with values in $[0, 1]$) to XYZ is the following:
\begin{equation}
    \begin{bmatrix}X\\Y\\Z\end{bmatrix}=\begin{bmatrix}0.4124564 & 0.3575761 & 0.1804375\\0.2126729 & 0.7151522 & 0.0721750\\0.0193339 & 0.1191920 & 0.9503041\end{bmatrix}\begin{bmatrix}R\\G\\B\end{bmatrix}
\end{equation}
Once we have the colors represented in the XYZ color space, we need to convert to the CIELAB color space. The conversion is the following:
\begin{equation}
\begin{aligned}
    L^\star &= 116 f\left(\frac{Y}{Y_n}\right)-16\\
    a^\star &= 500 \left( f\left(\frac{X}{X_n}\right)-f\left(\frac{Y}{Y_n}\right)\right)\\
    b^\star &= 200 \left( f\left(\frac{Y}{Y_n}\right)-f\left(\frac{Z}{Z_n}\right)\right)\\
\end{aligned}
\end{equation}
where:
\begin{equation}
    f(t) = \begin{cases}
        \sqrt[3]{t} & \text{if} \quad t>\delta^3\\
        \frac{t}{3\delta^2}+\frac{4}{29} &\text{otherwise}
    \end{cases}
\end{equation}
with $\delta=\frac{6}{29}$. Under the Illuminant D65 white reference, we have $X_n=95.0489$, $Y_n=100$ and $Z_n=108.8840$.

\section{Modified DLR loss}
\label{appendix:dlr_loss}
The original DLR loss proposed in \cite{croce2020reliable} is formulated as follows:
\begin{equation}
    \label{eq:original_dlr}
    \mathrm{DLR}(\z,y)=-\frac{\z_y - \max\limits_{i\neq y}\z_i}{\z_{\pi_1}-\z_{\pi_3}}
\end{equation}
where $\z = f(\x)$ and $\pi$ is the ordering of the element of $\z$ in decreasing order.
If a sample $\x$ is correctly classified, we have $\mathrm{DLR}(\z,y)\in[-1, 0]$ and $\x$ is misclassified only if $\mathrm{DLR}(\z,y)>0$. Croce \etal also propose a variant for untargeted attacks with a targeted objective. In some cases, performing a targeted attack against each class proved to be more successful at finding untargeted adversarial examples than simply performing an untargeted attack. The targeted variant is:
\begin{equation}
    \label{eq:original_dlr_targeted}
    \mathrm{tDLR}(\z,y)=-\frac{\z_y - \z_t}{\z_{\pi_1}-(\z_{\pi_3} + \z_{\pi_4})/2}
\end{equation}
where $t$ is the target class. For this variant, we can have no guarantee as to what $\x$ is classified as, simply by looking at the value of $\mathrm{tDLR}$.

For our optimization problem, we need to have a loss that is negative only when the misclassification or targeted classification is achieved. This way, we can formulate the misclassification or targeted classification constraint as $g(x)<0$. To this end, we modify $\mathrm{DLR}$ by taking the negative as follows:
\begin{equation}
    \dlr\tag{\ref{eq:dlr}}
\end{equation}
For targeted attack, we modify $\mathrm{tDLR}$ as follows:
\begin{equation}
    \label{eq:targeted_dlr}
    \mathrm{tDLR^+}(\z,y)=\frac{\max\limits_{i\neq t}\z_i - \z_t}{\z_{\pi_1}-(\z_{\pi_3} + \z_{\pi_4})/2}
\end{equation}
With these modifications, the misclassification  and targeted classification constraints are respected only when $\mathrm{DLR^+}$ and $\mathrm{tDLR^+}$ are negative. Conversely, the constraints are violated when these losses are positive, hence the $^+$ superscript.

\section{Penalty functions}
\label{appendix:penalties}
The four penalty functions plotted in \autoref{fig:penalties} are taken from \cite{birgin2005numerical} and defined as follows:
\begin{equation}
    \mathrm{PHR}(y,\rho,\mu) = \frac{1}{2\rho}(\max\{0, \mu+\rho y\}^2 - \mu^2)
\end{equation}
\begin{equation}
    P_1(y,\rho,\mu) = \begin{cases}
        \mu y + \frac{1}{2}\rho y^2+\rho^2y^3 &\text{if}\quad y\geq 0\\
        \mu y + \frac{1}{2}\rho y^2  &\text{if}\quad -\frac{\mu}{\rho} \leq y \leq 0\\
        -\frac{1}{2\rho}\mu^2 &\text{if}\quad y \leq -\frac{\mu}{\rho}
    \end{cases}
\end{equation}
\begin{equation}
    P_2(y,\rho,\mu) = \begin{cases}
        \mu y + \mu \rho y^2+ \frac{1}{6}\rho^2y^3 &\text{if}\quad y\geq 0\\
        \frac{\mu y}{1 - \rho y} &\text{if}\quad y \leq 0
    \end{cases}
\end{equation}
\begin{equation}
    P_3(y,\rho,\mu) = \begin{cases}
        \mu y + \mu \rho y^2 &\text{if}\quad y\geq 0\\
        \frac{\mu y}{1 - \rho y} &\text{if}\quad y \leq 0
    \end{cases}
\end{equation}

\section{Choice of $\alpha$}

For the experiments, we change the value of $\alpha$ according to the number of iterations. In our attack, $\alpha$ is a smoothing parameter. However, too much smoothing can degrade the performance of the attack for lower numbers of iterations. Therefore, we recommend values between 0.5 and 0.9 for numbers of iterations between 100 and 1\,000, and to keep $\alpha=0.9$ for more than 1\,000 iterations. Since our attack aims to find minimal adversarial perturbations, lower numbers of iterations are not recommended. 

\section{Hyper-parameter $\epsilon$ values}
\label{appendix:epsilon}

\autoref{tab:init_epsilon} reports the different $\epsilon$ used for each distance function in our experiments.

\begin{table}[h]
    \centering
    \begin{tabular}{cc}
         Distance & $\epsilon$ \\
         \midrule[1pt]
         $\ell_1$ & 0.5 \\
         $\ell_2$ & 0.1 \\
         SSIM & $3\times 10^{-5}$ \\
         CIEDE2000 & 0.05 \\
         LPIPS & $1\times 10^{-3}$ \\
         \bottomrule[1pt]
    \end{tabular}
    \caption{Initial values of $\epsilon$ for each distance.}
    \label{tab:init_epsilon}
\end{table}

\section{Additional results with SSIM}

We also tested our ALMA attack with the SSIM \cite{wang2004image} for which, to the best of our knowledge, no gradient-based attack currently exists. The only related work on adversarial attacks and SSIM is a black-box method \cite{gragnaniello2019perceptual}.
SSIM is a similarity function, so identical images have a SSIM of 1. Therefore, we minimize the quantity $1-\mathrm{SSIM}$ and report this instead of the SSIM.
The SSIM metric is defined between two gray-level images. Several modifications exist for color images, however, we simply considered the average SSIM over the color channels. 
\Autoref{tab:cifar10_SSIM_full, tab:imagenet_SSIM_full} report the results for all models on CIFAR10 and ImageNet respectively.

\section{Detailed experimental results}
\label{appendix:detailed_results}

\Autoref{tab:mnist_l1_full, tab:mnist_l2_full, tab:cifar10_l1_full, tab:cifar10_l2_full, tab:cifar10_SSIM_full, tab:cifar10_ciede2000_full, tab:cifar10_LPIPS_full, tab:imagenet_l1_full, tab:imagenet_l2_full, tab:imagenet_SSIM_full, tab:imagenet_ciede2000_full, tab:imagenet_LPIPS_full} report the detailed results for each dataset, model and attack. Results from \Autoref{tab:mnist_results, tab:cifar10_imagenet_results} are calculated from these tables using the geometric mean over the models. For the CIFAR10 and ImageNet models, RN stands for ResNet and WRN for Wide ResNet. For ImageNet, the targeted variant of FAB is used in the experiments denoted by a T superscript (see \autoref{sec:experiments} for details).

\begin{table}
    \centering
    \resizebox{\columnwidth}{!}{
    \begin{tabular}{clcccc}
    Model & Attack & ASR (\%) & \makecell{Median\\$\ell_1$} & \makecell{forwards /\\backwards} \\
    \midrule[1pt]
    \multirow{8}{*}{SmallCNN}
         & EAD 9$\times$100 \cite{chen2018ead} & 100 & 8.41 & 870 / 439 \\
         & EAD 9$\times$1000 \cite{chen2018ead} & 100 & 7.90 & 3\,810 / 1\,909 \\
         & FAB $\ell_1$ 100 \cite{croce2020minimally} & 100 & 6.31 & 201 / 1\,000 \\
         & FAB $\ell_1$ 1000 \cite{croce2020minimally} & 100 & 6.20 & 2\,001 / 10\,000 \\
         & FMN $\ell_1$ 100 \cite{pintor2021fast} & 95.02 & 6.50 & 100 / 100 \\
         & FMN $\ell_1$ 1000 \cite{pintor2021fast} & 95.19 & 6.39 & 1\,000 / 1\,000 \\
         & ALMA $\ell_1$ 100 & 100 & 6.77 & 100 / 100\\
         & ALMA $\ell_1$ 1000 & 100 & 6.23 & 1\,000 / 1\,000\\
    \midrule
    \multirow{8}{*}{\makecell{SmallCNN\\DDN}}
         & EAD 9$\times$100 \cite{chen2018ead} & 100 & 17.41 & 990 / 499 \\
         & EAD 9$\times$1000 \cite{chen2018ead} & 100 & 16.35 & 5\,010 / 2\,509 \\
         & FAB $\ell_1$ 100 \cite{croce2020minimally} & 100 & 16.53 & 201 / 1\,000 \\
         & FAB $\ell_1$ 1000 \cite{croce2020minimally} & 100 & 15.42 & 2\,001 / 10\,000 \\
         & FMN $\ell_1$ 100 \cite{pintor2021fast} & 99.97 & 16.09 & 100 / 100 \\
         & FMN $\ell_1$ 1000 \cite{pintor2021fast} & 99.97 & 15.67 & 1\,000 / 1\,000 \\
         & ALMA $\ell_1$ 100 & 100 & 14.73 & 100 / 100\\
         & ALMA $\ell_1$ 1000 & 100 & 14.02 & 1\,000 / 1\,000\\
    \midrule
    \multirow{8}{*}{\makecell{SmallCNN\\TRADES}}
         & EAD 9$\times$100 \cite{chen2018ead}& 100 & 14.80 & 967 / 486 \\
         & EAD 9$\times$1000 \cite{chen2018ead} & 100 & 12.15 & 6\,409 / 3\,208 \\
         & FAB $\ell_1$ 100 \cite{croce2020minimally} & 99.22 & 36.60 & 201 / 1\,000 \\
         & FAB $\ell_1$ 1000 \cite{croce2020minimally} & 99.35 & 32.37 & 2\,001 / 10\,000 \\
         & FMN $\ell_1$ 100 \cite{pintor2021fast} & 50.30 & 42.02 & 100 / 100 \\
         & FMN $\ell_1$ 1000 \cite{pintor2021fast} & 98.30 & 8.28 & 1\,000 / 1\,000 \\
         & ALMA $\ell_1$ 100 & 100 & 6.16 & 100 / 100\\
         & ALMA $\ell_1$ 1000 & 100 & 5.32 & 1\,000 / 1\,000\\
    \midrule
    \multirow{8}{*}{\makecell{CROWN\\IBP}}
         & EAD 9$\times$100 \cite{chen2018ead} & 89.17 & 106.95 & 509 / 258 \\
         & EAD 9$\times$1000 \cite{chen2018ead} & 91.32 & 86.45 & 5\,210 / 2\,609 \\
         & FAB $\ell_1$ 100 \cite{croce2020minimally} & 99.99 & 147.79 & 201 / 1\,000 \\
         & FAB $\ell_1$ 1000 \cite{croce2020minimally} & 99.99 & 110.96 & 2\,001 / 10\,000 \\
         & FMN $\ell_1$ 100 \cite{pintor2021fast} & 49.89 & -- & 100 / 100 \\
         & FMN $\ell_1$ 1000 \cite{pintor2021fast} & 88.36 & 3.50 & 1\,000 / 1\,000 \\
         & ALMA $\ell_1$ 100 & 99.59 & 27.94 & 100 / 100\\
         & ALMA $\ell_1$ 1000 & 100 & 5.65 & 1\,000 / 1\,000\\
    \bottomrule[1pt]
    \end{tabular}}
    \caption{Performance of the $\ell_1$ attacks on the MNIST dataset for each model.}
    \label{tab:mnist_l1_full}
\end{table}

\begin{table}
    \centering
    \resizebox{\columnwidth}{!}{
    \begin{tabular}{clcccc}
    Model & Attack & ASR (\%) & \makecell{Median\\$\ell_2$} & \makecell{forwards /\\backwards} \\
    \midrule[1pt]
    \multirow{11}{*}{SmallCNN}
         & C\&W $\ell_2$ 9$\times$1000 \cite{carlini2017towards} & 99.98 & 1.35 & 9\,000 / 9\,000 \\
         & C\&W $\ell_2$ 9$\times$10\,000 \cite{carlini2017towards} & 99.77 & 1.35 & 90\,000 / 90\,000  \\
         & DDN 100 \cite{rony2019decoupling} & 100 & 1.39 & 100 / 100 \\
         & DDN 1000 \cite{rony2019decoupling} & 100 & 1.37 & 1\,000 / 1\,000 \\
         & FAB $\ell_2$ 100 \cite{croce2020minimally} & 100 & 1.37 & 201 / 1\,000 \\
         & FAB $\ell_2$ 1000 \cite{croce2020minimally} & 100 & 1.36 & 2\,001 / 10\,000 \\
         & FMN $\ell_2$ 100 \cite{pintor2021fast} & 82.04 & 1.53 & 100 / 100 \\
         & FMN $\ell_2$ 1000 \cite{pintor2021fast} & 96.61 & 1.39 & 1\,000 / 1\,000 \\
         & APGD$^\text{T}_\text{DLR}$ $\ell_2$ \cite{croce2020reliable} & 100 & 1.31 & 13\,082 / 13\,062 \\
         & ALMA $\ell_2$ 100 & 100 & 1.38 & 100 / 100\\
         & ALMA $\ell_2$ 1000 & 100 & 1.32 & 1\,000 / 1\,000\\
    \midrule
    \multirow{11}{*}{\makecell{SmallCNN\\DDN}}
         & C\&W $\ell_2$ 9$\times$1000 \cite{carlini2017towards} & 99.96 & 2.76 & 9\,000 / 9\,000 \\
         & C\&W $\ell_2$ 9$\times$10\,000 \cite{carlini2017towards} & 99.59 & 2.69 & 90\,000 / 90\,000  \\
         & DDN 100 \cite{rony2019decoupling} & 100 & 2.74 & 100 / 100 \\
         & DDN 1000 \cite{rony2019decoupling} & 100 & 2.66 & 1\,000 / 1\,000 \\
         & FAB $\ell_2$ 100 \cite{croce2020minimally} & 100 & 2.74 & 201 / 1\,000 \\
         & FAB $\ell_2$ 1000 \cite{croce2020minimally} & 100 & 2.71 & 2\,001 / 10\,000 \\
         & FMN $\ell_2$ 100 \cite{pintor2021fast} & 99.95 & 2.67 & 100 / 100 \\
         & FMN $\ell_2$ 1000 \cite{pintor2021fast} & 100 & 2.67 & 1\,000 / 1\,000 \\
         & APGD$^\text{T}_\text{DLR}$ $\ell_2$ \cite{croce2020reliable} & 100 & 2.58 & 13\,410 / 13\,390 \\
         & ALMA $\ell_2$ 100 & 100 & 2.68 & 100 / 100\\
         & ALMA $\ell_2$ 1000 & 100 & 2.59 & 1\,000 / 1\,000\\
    \midrule
    \multirow{11}{*}{\makecell{SmallCNN\\TRADES}}
         & C\&W $\ell_2$ 9$\times$1000 \cite{carlini2017towards} & 99.99 & 3.32 & 9\,000 / 9\,000 \\
         & C\&W $\ell_2$ 9$\times$10\,000 \cite{carlini2017towards} & 99.97 & 2.28 & 90\,000 / 90\,000  \\
         & DDN 100 \cite{rony2019decoupling} & 99.69 & 2.17 & 100 / 100 \\
         & DDN 1000 \cite{rony2019decoupling} & 100 & 1.91 & 1\,000 / 1\,000 \\
         & FAB $\ell_2$ 100 \cite{croce2020minimally} & 99.88 & 1.77 & 201 / 1\,000 \\
         & FAB $\ell_2$ 1000 \cite{croce2020minimally} & 99.90 & 1.74 & 2\,001 / 10\,000 \\
         & FMN $\ell_2$ 100 \cite{pintor2021fast} & 86.41 & 2.24 & 100 / 100 \\
         & FMN $\ell_2$ 1000 \cite{pintor2021fast} & 99.83 & 1.99 & 1\,000 / 1\,000 \\
         & APGD$^\text{T}_\text{DLR}$ $\ell_2$ \cite{croce2020reliable} & 100 & 3.36 & 13\,919 / 13\,899 \\
         & ALMA $\ell_2$ 100 & 100 & 1.74 & 100 / 100\\
         & ALMA $\ell_2$ 1000 & 100 & 1.55 & 1\,000 / 1\,000\\
    \midrule
    \multirow{11}{*}{\makecell{CROWN\\IBP}}
         & C\&W $\ell_2$ 9$\times$1000 \cite{carlini2017towards} & 2.61 & -- & 9\,000 / 9\,000 \\
         & C\&W $\ell_2$ 9$\times$10\,000 \cite{carlini2017towards} & 2.63 & -- & 90\,000 / 90\,000  \\
         & DDN 100 \cite{rony2019decoupling} & 94.34 & 1.46 & 100 / 100 \\
         & DDN 1000 \cite{rony2019decoupling} & 99.27 & 0.97 & 1\,000 / 1\,000 \\
         & FAB $\ell_2$ 100 \cite{croce2020minimally} & 99.98 & 5.19 & 201 / 1\,000 \\
         & FAB $\ell_2$ 1000 \cite{croce2020minimally} & 99.98 & 3.34 & 2\,001 / 10\,000 \\
         & FMN $\ell_2$ 100 \cite{pintor2021fast} & 67.80 & 2.14 & 100 / 100 \\
         & FMN $\ell_2$ 1000 \cite{pintor2021fast} & 89.08 & 1.34 & 1\,000 / 1\,000 \\
         & APGD$^\text{T}_\text{DLR}$ $\ell_2$ \cite{croce2020reliable} & 99.94 & 3.57 & 9\,286 / 9\,273 \\
         & ALMA $\ell_2$ 100 & 98.90 & 4.96 & 100 / 100\\
         & ALMA $\ell_2$ 1000 & 100 & 1.26 & 1\,000 / 1\,000\\
    \bottomrule[1pt]
    \end{tabular}}
    \caption{Performance of the $\ell_2$ attacks on the MNIST dataset for each model.}
    \label{tab:mnist_l2_full}
\end{table}

\begin{table}
    \centering
    \resizebox{\columnwidth}{!}{
    \begin{tabular}{clcccc}
    Model & Attack & ASR (\%) & \makecell{Median\\$\ell_1$} & \makecell{forwards /\\backwards} \\
    \midrule[1pt]
    \multirow{8}{*}{\makecell{WRN\\28-10}}
         & EAD 9$\times$100 \cite{chen2018ead} & 100 & 1.79 & 530 / 269 \\
         & EAD 9$\times$1000 \cite{chen2018ead} & 100 & 1.62 & 4\,910 / 2\,459 \\
         & FAB $\ell_1$ 100 \cite{croce2020minimally} & 92.3 & 1.27 & 200 / 1\,000 \\
         & FAB $\ell_1$ 1000 \cite{croce2020minimally} & 98.8 & 1.07 & 2\,000 / 10\,000 \\
         & FMN $\ell_1$ 100 \cite{pintor2021fast} & 99.7 & 1.01 & 100 / 100 \\
         & FMN $\ell_1$ 1000 \cite{pintor2021fast} & 99.5 & 0.98 & 1\,000 / 1\,000 \\
         & ALMA $\ell_1$ 100 & 100 & 1.26 & 100 / 100\\
         & ALMA $\ell_1$ 1000 & 100 & 1.02 & 1\,000 / 1\,000\\
    \midrule
    \multirow{8}{*}{\makecell{WRN\\28-10\\Carmon\\\etal \cite{carmon2019unlabeled}}}
         & EAD 9$\times$100 \cite{chen2018ead} & 100 & 6.62 & 600 / 304 \\
         & EAD 9$\times$1000 \cite{chen2018ead} & 100 & 6.07 & 3\,760 / 1\,884 \\
         & FAB $\ell_1$ 100 \cite{croce2020minimally} & 97.8 & 5.57 & 200 / 1\,000 \\
         & FAB $\ell_1$ 1000 \cite{croce2020minimally} & 98.2 & 5.07 & 2\,000 / 10\,000 \\
         & FMN $\ell_1$ 100 \cite{pintor2021fast} & 100 & 4.70 & 100 / 100 \\
         & FMN $\ell_1$ 1000 \cite{pintor2021fast} & 100 & 4.64 & 1\,000 / 1\,000 \\
         & ALMA $\ell_1$ 100 & 100 & 5.20 & 100 / 100\\
         & ALMA $\ell_1$ 1000 & 100 & 4.75 & 1\,000 / 1\,000\\
    \midrule
    \multirow{8}{*}{\makecell{RN-50\\Augustin\\\etal \cite{augustin2020adversarial}}} 
         & EAD 9$\times$100 \cite{chen2018ead} & 100 & 19.18 & 590 / 299 \\
         & EAD 9$\times$1000 \cite{chen2018ead} & 100 & 16.39 & 4\,260 / 2\,134 \\
         & FAB $\ell_1$ 100 \cite{croce2020minimally} & 99.8 & 10.95 & 200 / 1\,000 \\
         & FAB $\ell_1$ 1000 \cite{croce2020minimally} & 99.8 & 10.09 & 2\,000 / 10\,000 \\
         & FMN $\ell_1$ 100 \cite{pintor2021fast} & 100 & 10.21 & 100 / 100 \\
         & FMN $\ell_1$ 1000 \cite{pintor2021fast} & 100 & 9.79 & 1\,000 / 1\,000 \\
         & ALMA $\ell_1$ 100 & 100 & 12.15 & 100 / 100\\
         & ALMA $\ell_1$ 1000 & 100 & 10.35 & 1\,000 / 1\,000\\
    \bottomrule[1pt]
    \end{tabular}}
    \caption{Performance of the $\ell_1$ attacks on the CIFAR10 dataset for each model.}
    \label{tab:cifar10_l1_full}
\end{table}

\begin{table}
    \centering
    \resizebox{\columnwidth}{!}{
    \begin{tabular}{clcccc}
    Model & Attack & ASR (\%) & \makecell{Median\\$\ell_2$} & \makecell{forwards /\\backwards} \\
    \midrule[1pt]
    \multirow{11}{*}{\makecell{WRN\\28-10}} 
         & C\&W $\ell_2$ 9$\times$1000 \cite{carlini2017towards} & 100 & 0.10 & 9\,000 / 9\,000 \\
         & C\&W $\ell_2$ 9$\times$10\,000 \cite{carlini2017towards} & 100 & 0.10 & 90\,000 / 90\,000  \\
         & DDN 100 \cite{rony2019decoupling} & 100 & 0.11 & 100 / 100 \\
         & DDN 1000 \cite{rony2019decoupling} & 100 & 0.11 & 1\,000 / 1\,000 \\
         & FAB $\ell_2$ 100 \cite{croce2020minimally} & 100 & 0.09 & 201 / 1\,000 \\
         & FAB $\ell_2$ 1000 \cite{croce2020minimally} & 100 & 0.09 & 2\,001 / 10\,000 \\
         & FMN $\ell_2$ 100 \cite{pintor2021fast} & 99.7 & 0.12 & 100 / 100 \\
         & FMN $\ell_2$ 1000 \cite{pintor2021fast} & 99.5 & 0.09 & 1\,000 / 1\,000 \\
         & APGD$^\text{T}_\text{DLR}$ $\ell_2$ \cite{croce2020reliable} & 100 & 0.09 & 4\,336 / 4\,312 \\
         & ALMA $\ell_2$ 100 & 100 & 0.09 & 100 / 100\\
         & ALMA $\ell_2$ 1000 & 100 & 0.09 & 1\,000 / 1\,000\\
    \midrule
    \multirow{11}{*}{\makecell{WRN\\28-10\\Carmon\\\etal \cite{carmon2019unlabeled}}}
         & C\&W $\ell_2$ 9$\times$1000 \cite{carlini2017towards} & 100 & 0.70 & 7\,502 / 7\,500 \\
         & C\&W $\ell_2$ 9$\times$10\,000 \cite{carlini2017towards} & 100 & 0.70 & 71\,602 / 71\,600  \\
         & DDN 100 \cite{rony2019decoupling} & 100 & 0.72 & 100 / 100 \\
         & DDN 1000 \cite{rony2019decoupling} & 100 & 0.71 & 1\,000 / 1\,000 \\
         & FAB $\ell_2$ 100 \cite{croce2020minimally} & 100 & 0.71 & 201 / 1\,000 \\
         & FAB $\ell_2$ 1000 \cite{croce2020minimally} & 100 & 0.71 & 2\,001 / 10\,000 \\
         & FMN $\ell_2$ 100 \cite{pintor2021fast} & 100 & 0.69 & 100 / 100 \\
         & FMN $\ell_2$ 1000 \cite{pintor2021fast} & 100 & 0.70 & 1\,000 / 1\,000 \\
         & APGD$^\text{T}_\text{DLR}$ $\ell_2$ \cite{croce2020reliable} & 100 & 0.68 & 5\,683 / 5\,659 \\
         & ALMA $\ell_2$ 100 & 100 & 0.70 & 100 / 100\\
         & ALMA $\ell_2$ 1000 & 100 & 0.67 & 1\,000 / 1\,000\\
    \midrule
    \multirow{11}{*}{\makecell{RN-50\\Augustin\\\etal \cite{augustin2020adversarial}}} 
         & C\&W $\ell_2$ 9$\times$1000 \cite{carlini2017towards} & 100 & 0.96 & 7\,515 / 7\,513 \\
         & C\&W $\ell_2$ 9$\times$10\,000 \cite{carlini2017towards} & 100 & 0.95 & 73\,869 / 73\,867  \\
         & DDN 100 \cite{rony2019decoupling} & 100 & 0.97 & 100 / 100 \\
         & DDN 1000 \cite{rony2019decoupling} & 100 & 0.96 & 1\,000 / 1\,000 \\
         & FAB $\ell_2$ 100 \cite{croce2020minimally} & 100 & 1.01 & 201 / 1\,000 \\
         & FAB $\ell_2$ 1000 \cite{croce2020minimally} & 100 & 1.00 & 2\,001 / 10\,000 \\
         & FMN $\ell_2$ 100 \cite{pintor2021fast} & 100 & 0.95 & 100 / 100 \\
         & FMN $\ell_2$ 1000 \cite{pintor2021fast} & 100 & 0.96 & 1\,000 / 1\,000 \\
         & APGD$^\text{T}_\text{DLR}$ $\ell_2$ \cite{croce2020reliable} & 100 & 0.91 & 6\,198 / 6\,174 \\
         & ALMA $\ell_2$ 100 & 100 & 0.98 & 100 / 100\\
         & ALMA $\ell_2$ 1000 & 100 & 0.92 & 1\,000 / 1\,000\\
    \bottomrule[1pt]
    \end{tabular}}
    \caption{Performance of the $\ell_2$ attacks on the CIFAR10 dataset for each model.}
    \label{tab:cifar10_l2_full}
\end{table}

\begin{table}
    \centering
    \resizebox{\columnwidth}{!}{
    \begin{tabular}{clcccc}
    Model & Attack & ASR (\%) & \makecell{Median\\CIEDE2000} & \makecell{forwards /\\backwards} \\
    \midrule[1pt]
    \multirow{5}{*}{\makecell{WRN\\28-10}}
         & C\&W CIEDE2000 9$\times$1000 & 100 & 0.23 & 7\,741 / 7\,740 \\
         & Perc-AL 100 \cite{zhao2020towards} & 100 & 0.86 & 201 / 100 \\
         & Perc-AL 1000 \cite{zhao2020towards} & 100 & 0.72 & 2\,001 / 1\,000 \\
         & ALMA CIEDE2000 100 & 100 & 0.18 & 100 / 100\\
         & ALMA CIEDE2000 1000 & 100 & 0.14 & 1\,000 / 1\,000\\
    \midrule
    \multirow{5}{*}{\makecell{WRN\\28-10\\Carmon\\\etal \cite{carmon2019unlabeled}}}
         & C\&W CIEDE2000 9$\times$1000 & 100 & 2.12 & 6\,243 / 6\,240 \\
         & Perc-AL 100 \cite{zhao2020towards} & 100 & 5.69 & 201 / 100 \\
         & Perc-AL 1000 \cite{zhao2020towards} & 100 & 5.82 & 2\,001 / 1\,000 \\
         & ALMA CIEDE2000 100 & 100 & 3.65 & 100 / 100\\
         & ALMA CIEDE2000 1000 & 100 & 2.08 & 1\,000 / 1\,000\\
    \midrule
    \multirow{5}{*}{\makecell{RN-50\\Augustin\\\etal \cite{augustin2020adversarial}}}
         & C\&W CIEDE2000 9$\times$1000 & 100 & 1.63 & 6\,303 / 6\,300 \\
         & Perc-AL 100 \cite{zhao2020towards} & 100 & 4.85 & 201 / 100 \\
         & Perc-AL 1000 \cite{zhao2020towards} & 100 & 4.83 & 2\,001 / 1\,000 \\
         & ALMA CIEDE2000 100 & 99.8 & 1.94 & 100 / 100\\
         & ALMA CIEDE2000 1000 & 99.8 & 1.58 & 1\,000 / 1\,000\\
    \bottomrule[1pt]
    \end{tabular}}
    \caption{Performance of the CIEDE2000 attacks on the CIFAR10 dataset for each model.}
    \label{tab:cifar10_ciede2000_full}
\end{table}

\begin{table}
    \centering
    \resizebox{\columnwidth}{!}{
    \begin{tabular}{clcccc}
    Model & Attack & ASR (\%) & \makecell{Median\\LPIPS\\$\times 10^{-2}$} & \makecell{forwards /\\backwards} \\
    \midrule[1pt]
    \multirow{4}{*}{\makecell{WRN 28-10}} 
         & C\&W LPIPS 9$\times$1000 & 100 & 0.32 & 4\,565 / 4\,560 \\
         & LPA\textsuperscript{\ddag} \cite{laidlaw2021perceptual} & 100 & 4.81 & 1\,129 / 1\,119 \\
         & ALMA LPIPS 100 & 100 & 0.29 & 100 / 100\\
         & ALMA LPIPS 1000 & 100 & 0.12 & 1\,000 / 1\,000\\
    \midrule
    \multirow{4}{*}{\makecell{WRN 28-10\\Carmon \etal \cite{carmon2019unlabeled}}}
         & C\&W LPIPS 9$\times$1000 & 100 & 0.50 & 7\,981 / 7\,980 \\
         & LPA\textsuperscript{\ddag} \cite{laidlaw2021perceptual} & 100 & 5.06 & 1\,092 / 1\,082 \\
         & ALMA LPIPS 100 & 100 & 6.76 & 100 / 100\\
         & ALMA LPIPS 1000 & 100 & 1.01 & 1\,000 / 1\,000\\
    \midrule
    \multirow{4}{*}{\makecell{RN-50\\Augustin \etal \cite{augustin2020adversarial}}}
         & C\&W LPIPS 9$\times$1000 & 100 & 0.64 & 8\,101 / 8\,100 \\
         & LPA\textsuperscript{\ddag} \cite{laidlaw2021perceptual} & 100 & 6.42 & 1\,133 / 1\,123 \\
         & ALMA LPIPS 100 & 99.9 & 7.66 & 100 / 100\\
         & ALMA LPIPS 1000 & 100 & 1.82 & 1\,000 / 1\,000\\
    \bottomrule[1pt]
    \end{tabular}}
    \caption{Performance of the LPIPS variant of ALMA on the CIFAR10 dataset for each model. \textsuperscript{\ddag}A binary search is performed on each sample to get a minimal perturbation attack (\autoref{eq:minimal_adv_optim}).}
    \label{tab:cifar10_LPIPS_full}
\end{table}

\begin{table}
    \centering
    \resizebox{\columnwidth}{!}{
    \begin{tabular}{clcccc}
    Model & Attack & ASR (\%) & \makecell{Median\\$1-$SSIM\\$\times 10^{-4}$} & \makecell{forwards /\\backwards} \\
    \midrule[1pt]
    \multirow{2}{*}{\makecell{WRN\\28-10}} 
         & ALMA SSIM 100 & 100 & 0.4 & 100 / 100\\
         & ALMA SSIM 1000 & 100 & 0.1 & 1\,000 / 1\,000\\
    \midrule
    \multirow{2}{*}{\makecell{Wide ResNet 28-10\\Carmon \etal \cite{carmon2019unlabeled}}}
         & ALMA SSIM 100 & 100 & 7.5 & 100 / 100\\
         & ALMA SSIM 1000 & 100 & 2.8 & 1\,000 / 1\,000\\
    \midrule
    \multirow{2}{*}{\makecell{ResNet-50\\Augustin \etal \cite{augustin2020adversarial}}}
         & ALMA SSIM 100 & 100 & 4.1 & 100 / 100\\
         & ALMA SSIM 1000 & 100 & 2.0 & 1\,000 / 1\,000\\
    \bottomrule[1pt]
    \end{tabular}}
    \caption{Performance of the SSIM variant of ALMA on the CIFAR10 dataset for each model.}
    \label{tab:cifar10_SSIM_full}
\end{table}

\begin{table}
    \centering
    \resizebox{\columnwidth}{!}{
    \begin{tabular}{clcccc}
    Model & Attack & ASR (\%) & \makecell{Median\\$\ell_1$} & \makecell{forwards /\\backwards} \\
    \midrule[1pt]
    \multirow{8}{*}{RN-50} 
         & EAD 9$\times$100 \cite{chen2018ead} & 100 & 6.70 & 437 / 222 \\
         & EAD 9$\times$1000 \cite{chen2018ead} & 100 & 6.08 & 4\,510 / 2\,259 \\
         & FAB$^\text{T}$ $\ell_1$ 100 \cite{croce2020minimally} & 74.4 & 9.01 & 1\,810 / 900 \\
         & FAB$^\text{T}$ $\ell_1$ 1000 \cite{croce2020minimally} & 81.0 & 4.82 & 18\,010 / 9\,000 \\
         & FMN $\ell_1$ 100 \cite{pintor2021fast} & 95.6 & 3.72 & 100 / 100 \\
         & FMN $\ell_1$ 1000 \cite{pintor2021fast} & 94.5 & 3.43 & 1\,000 / 1\,000 \\
         & ALMA $\ell_1$ 100 & 100 & 8.47 & 100 / 100\\
         & ALMA $\ell_1$ 1000 & 100 & 4.25 & 1\,000 / 1\,000\\
    \midrule
    \multirow{8}{*}{\makecell{RN-50\\$\ell_2$-AT}}
         & EAD 9$\times$100 \cite{chen2018ead} & 100 & 62.21 & 458 / 233 \\
         & EAD 9$\times$1000 \cite{chen2018ead} & 100 & 55.16 & 3\,450 / 1\,729 \\
         & FAB$^\text{T}$ $\ell_1$ 100 \cite{croce2020minimally} & 98.5 & 31.33 & 1\,810 / 900 \\
         & FAB$^\text{T}$ $\ell_1$ 1000 \cite{croce2020minimally} & 93.2 & 33.58 & 18\,010 / 9\,000 \\
         & FMN $\ell_1$ 100 \cite{pintor2021fast} & 100 & 36.68 & 100 / 100 \\
         & FMN $\ell_1$ 1000 \cite{pintor2021fast} & 100 & 30.52 & 1\,000 / 1\,000 \\
         & ALMA $\ell_1$ 100 & 100 & 61.37 & 100 / 100\\
         & ALMA $\ell_1$ 1000 & 100 & 40.41 & 1\,000 / 1\,000\\
    \midrule
    \multirow{8}{*}{\makecell{RN-50\\$\ell_\infty$-AT}} 
         & EAD 9$\times$100 \cite{chen2018ead} & 100 & 6.40 & 582 / 295 \\
         & EAD 9$\times$1000 \cite{chen2018ead} & 100 & 6.29 & 3\,410 / 1\,709 \\
         & FAB$^\text{T}$ $\ell_1$ 100 \cite{croce2020minimally} & 95.6 & 4.36 & 1\,810 / 900 \\
         & FAB$^\text{T}$ $\ell_1$ 1000 \cite{croce2020minimally} & 93.6 & 4.33 & 18\,010 / 9\,000 \\
         & FMN $\ell_1$ 100 \cite{pintor2021fast} & 87.8 & 4.16 & 100 / 100 \\
         & FMN $\ell_1$ 1000 \cite{pintor2021fast} & 87.7 & 4.16 & 1\,000 / 1\,000 \\
         & ALMA $\ell_1$ 100 & 100 & 14.90 & 100 / 100\\
         & ALMA $\ell_1$ 1000 & 100 & 10.33 & 1\,000 / 1\,000\\
    \bottomrule[1pt]
    \end{tabular}}
    \caption{Performance of the $\ell_1$ attacks on the ImageNet dataset for each model.}
    \label{tab:imagenet_l1_full}
\end{table}

\begin{table}
    \centering
    \resizebox{\columnwidth}{!}{
    \begin{tabular}{clcccc}
    Model & Attack & ASR (\%) & \makecell{Median\\$\ell_2$} & \makecell{forwards /\\backwards} \\
    \midrule[1pt]
    \multirow{11}{*}{RN-50} 
         & C\&W $\ell_2$ 9$\times$1000 \cite{carlini2017towards} & 100 & 0.21 & 8\,775 / 8\,775 \\
         & C\&W $\ell_2$ 9$\times$10\,000 \cite{carlini2017towards} & 100 & 0.21 & 82\,668 / 82\,667  \\
         & DDN 100 \cite{rony2019decoupling} & 99.8 & 0.18 & 100 / 100 \\
         & DDN 1000 \cite{rony2019decoupling} & 99.9 & 0.17 & 1\,000 / 1\,000 \\
         & FAB$^\text{T}$ $\ell_2$ 100 \cite{croce2020minimally} & 99.3 & 0.10 & 1\,810 / 900 \\
         & FAB$^\text{T}$ $\ell_2$ 1000 \cite{croce2020minimally} & 98.0 & 0.10 & 18\,010 / 9\,000 \\
         & FMN $\ell_2$ 100 \cite{pintor2021fast} & 98.9 & 0.12 & 100 / 100 \\
         & FMN $\ell_2$ 1000 \cite{pintor2021fast} & 99.3 & 0.10 & 1\,000 / 1\,000 \\
         & APGD$^\text{T}_\text{DLR}$ $\ell_2$ \cite{croce2020reliable} & 100 & 0.09 & 4\,866 \ 4\,838 \\
         & ALMA $\ell_2$ 100 & 100 & 0.10 & 100 / 100\\
         & ALMA $\ell_2$ 1000 & 100 & 0.10 & 1000 / 1000\\
    \midrule
    \multirow{11}{*}{\makecell{RN-50\\$\ell_2$-AT}}
         & C\&W $\ell_2$ 9$\times$1000 \cite{carlini2017towards} & 99.9 & 1.17 & 6\,260 / 6\,256 \\
         & C\&W $\ell_2$ 9$\times$10\,000 \cite{carlini2017towards} & 99.9 & 1.17 & 57\,004 / 52\,000  \\
         & DDN 100 \cite{rony2019decoupling} & 99.5 & 1.09 & 100 / 100 \\
         & DDN 1000 \cite{rony2019decoupling} & 99.7 & 1.10 & 1\,000 / 1\,000 \\
         & FAB$^\text{T}$ $\ell_2$ 100 \cite{croce2020minimally} & 100 & 0.81 & 1\,810 / 900 \\
         & FAB$^\text{T}$ $\ell_2$ 1000 \cite{croce2020minimally} & 99.3 & 0.81 & 18\,010 / 9\,000 \\
         & FMN $\ell_2$ 100 \cite{pintor2021fast} & 99.6 & 0.84 & 100 / 100 \\
         & FMN $\ell_2$ 1000 \cite{pintor2021fast} & 99.9 & 0.82 & 1\,000 / 1\,000 \\
         & APGD$^\text{T}_\text{DLR}$ $\ell_2$ \cite{croce2020reliable} & 100 & 0.80 & 7\,005 / 6\,977 \\
         & ALMA $\ell_2$ 100 & 100 & 0.85 & 100 / 100\\
         & ALMA $\ell_2$ 1000 & 100 & 0.84 & 1\,000 / 1\,000\\
    \midrule
    \multirow{11}{*}{\makecell{RN-50\\$\ell_\infty$-AT}} 
         & C\&W $\ell_2$ 9$\times$1000 \cite{carlini2017towards} & 99.6 & 0.76 & 6\,933 / 6\,930 \\
         & C\&W $\ell_2$ 9$\times$10\,000 \cite{carlini2017towards} & 99.6 & 0.76 & 65\,203 / 65\,200  \\
         & DDN 100 \cite{rony2019decoupling} & 99.8 & 0.67 & 100 / 100 \\
         & DDN 1000 \cite{rony2019decoupling} & 100 & 0.66 & 1\,000 / 1\,000 \\
         & FAB$^\text{T}$ $\ell_2$ 100 \cite{croce2020minimally} & 99.8 & 0.55 & 1\,810 / 900 \\
         & FAB$^\text{T}$ $\ell_2$ 1000 \cite{croce2020minimally} & 99.4 & 0.55 & 18\,010 / 9\,000 \\
         & FMN $\ell_2$ 100 \cite{pintor2021fast} & 99.8 & 0.57 & 100 / 100 \\
         & FMN $\ell_2$ 1000 \cite{pintor2021fast} & 99.7 & 0.57 & 1\,000 / 1\,000 \\
         & APGD$^\text{T}_\text{DLR}$ $\ell_2$ \cite{croce2020reliable} & 100 & 0.54 & 6\,647 / 6\,619 \\
         & ALMA $\ell_2$ 100 & 100 & 0.62 & 100 / 100\\
         & ALMA $\ell_2$ 1000 & 100 & 0.54 & 1\,000 / 1\,000\\
    \bottomrule[1pt]
    \end{tabular}}
    \caption{Performance of the $\ell_2$ attacks on the ImageNet dataset for each model.}
    \label{tab:imagenet_l2_full}
\end{table}

\begin{table}
    \centering
    \resizebox{\columnwidth}{!}{
    \begin{tabular}{clcccc}
    Model & Attack & ASR (\%) & \makecell{Median\\CIEDE2000} & \makecell{forwards /\\backwards} \\
    \midrule[1pt]
    \multirow{5}{*}{RN-50} 
         & C\&W CIEDE2000 9$\times$1000 & 100 & 0.80 & 4\,505 / 4\,500 \\
         & Perc-AL 100 \cite{zhao2020towards} & 100 & 1.31 & 2\,01 / 100 \\
         & Perc-AL 1000 \cite{zhao2020towards} & 100 & 1.07 & 2\,001 / 1\,000 \\
         & ALMA CIEDE2000 100 & 100 & 0.17 & 100 / 100\\
         & ALMA CIEDE2000 1000 & 100 & 0.13 & 1\,000 / 1\,000\\
    \midrule
    \multirow{5}{*}{\makecell{RN-50\\$\ell_2$-AT}}
         & C\&W CIEDE2000 9$\times$1000 & 100 & 1.64 & 6\,303 / 6\,300 \\
         & Perc-AL 100 \cite{zhao2020towards} & 99.9 & 5.87 & 201 / 100 \\
         & Perc-AL 1000 \cite{zhao2020towards} & 99.9 & 6.07 & 2\,001 / 1\,000 \\
         & ALMA CIEDE2000 100 & 100 & 1.51 & 100 / 100\\
         & ALMA CIEDE2000 1000 & 100 & 1.34 & 1\,000 / 1\,000\\
    \midrule
    \multirow{5}{*}{\makecell{RN-50\\$\ell_\infty$-AT}}
         & C\&W CIEDE2000 9$\times$1000 & 100 & 2.05 & 6\,303 / 6\,300 \\
         & Perc-AL 100 \cite{zhao2020towards} & 99.8 & 5.82 & 201 / 100 \\
         & Perc-AL 1000 \cite{zhao2020towards} & 99.9 & 6.12 & 2\,001 / 1\,000 \\
         & ALMA CIEDE2000 100 & 100 & 1.61 & 100 / 100\\
         & ALMA CIEDE2000 1000 & 100 & 1.46 & 1\,000 / 1\,000\\
    \bottomrule[1pt]
    \end{tabular}}
    \caption{Performance of the CIEDE2000 attacks on the CIFAR10 dataset for each model.}
    \label{tab:imagenet_ciede2000_full}
\end{table}

\begin{table}
    \centering
    \resizebox{\columnwidth}{!}{
    \begin{tabular}{clcccc}
    Model & Attack & ASR (\%) & \makecell{Median\\LPIPS\\$\times 10^{-2}$} & \makecell{forwards /\\backwards} \\
    \midrule[1pt]
    \multirow{4}{*}{RN-50} 
         & C\&W LPIPS 9$\times$1000 & 100 & 2.39 & 2\,332 / 2\,325 \\
         & LPA\textsuperscript{\ddag} \cite{laidlaw2021perceptual} & 100 & 5.02 & 1\,159 / 1\,149 \\
         & ALMA LPIPS 100 & 100 & 0.34 & 100 / 100\\
         & ALMA LPIPS 1000 & 100 & 0.24 & 1\,000 / 1\,000\\
    \midrule
    \multirow{4}{*}{\makecell{RN-50\\$\ell_2$-AT}}
         & C\&W LPIPS 9$\times$1000 & 100 & 2.22 & 7\,701 / 7\,700 \\
         & LPA\textsuperscript{\ddag} \cite{laidlaw2021perceptual} & 100 & 6.83 & 1\,257 / 1\,247 \\
         & ALMA LPIPS 100 & 100 & 3.96 & 100 / 100\\
         & ALMA LPIPS 1000 & 100 & 2.90 & 1\,000 / 1\,000\\
    \midrule
    \multirow{4}{*}{\makecell{RN-50\\$\ell_\infty$-AT}}
         & C\&W LPIPS 9$\times$1000 & 100 & 1.68 & 6\,753 / 6\,750 \\
         & LPA\textsuperscript{\ddag} \cite{laidlaw2021perceptual} & 100 & 5.66 & 1\,218 / 1\,208 \\
         & ALMA LPIPS 100 & 100 & 2.99 & 100 / 100\\
         & ALMA LPIPS 1000 & 100 & 2.11 & 1\,000 / 1\,000\\
    \bottomrule[1pt]
    \end{tabular}}
    \caption{Performance of the LPIPS variant of ALMA on the ImageNet dataset for each model. \textsuperscript{\ddag}A binary search is performed on each sample to get a minimal perturbation attack (\autoref{eq:minimal_adv_optim}).}
    \label{tab:imagenet_LPIPS_full}
\end{table}

\begin{table}
    \centering
    \resizebox{\columnwidth}{!}{
    \begin{tabular}{clcccc}
    Model & Attack & ASR (\%) & \makecell{Median\\$1-$SSIM\\$\times 10^{-5}$} & \makecell{forwards /\\backwards} \\
    \midrule[1pt]
    \multirow{2}{*}{RN-50} 
         & ALMA SSIM 100 & 100 & 0.44 & 100 / 100\\
         & ALMA SSIM 1000 & 100 & 0.05 & 1\,000 / 1\,000\\
    \midrule
    \multirow{2}{*}{\makecell{RN-50\\$\ell_2$-AT}}
         & ALMA SSIM 100 & 100 & 16.13 & 100 / 100\\
         & ALMA SSIM 1000 & 100 & 5.58 & 1\,000 / 1\,000\\
    \midrule
    \multirow{2}{*}{\makecell{RN-50\\$\ell_\infty$-AT}}
         & ALMA SSIM 100 & 100 & 8.69 & 100 / 100\\
         & ALMA SSIM 1000 & 100 & 2.77 & 1\,000 / 1\,000\\
    \bottomrule[1pt]
    \end{tabular}}
    \caption{Performance of the SSIM variant of ALMA on the ImageNet dataset for each model.}
    \label{tab:imagenet_SSIM_full}
\end{table}

\section{Robust Accuracy curves}
\label{appendix:curves}

\begin{figure}
    \centering
    \subfloat[SmallCNN]{\includegraphics[width=0.7\columnwidth]{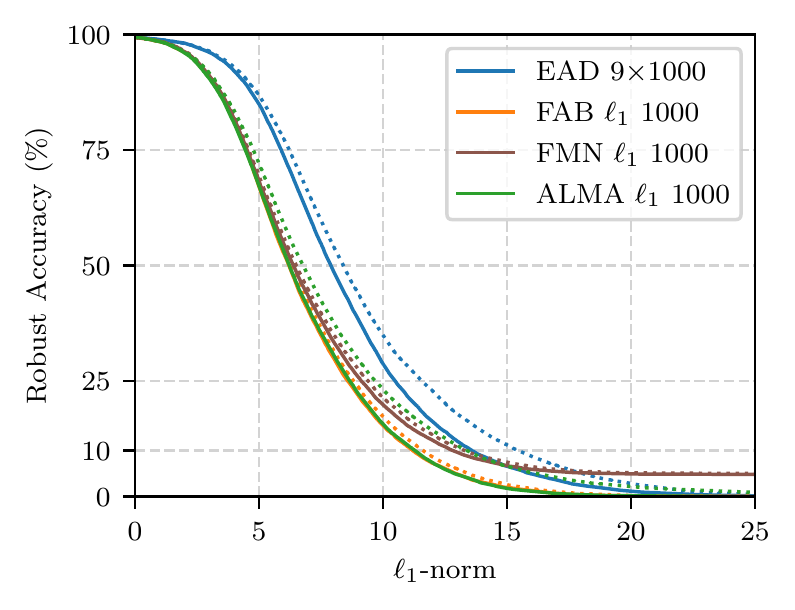}}\\
    \subfloat[SmallCNN-DDN]{\includegraphics[width=0.7\columnwidth]{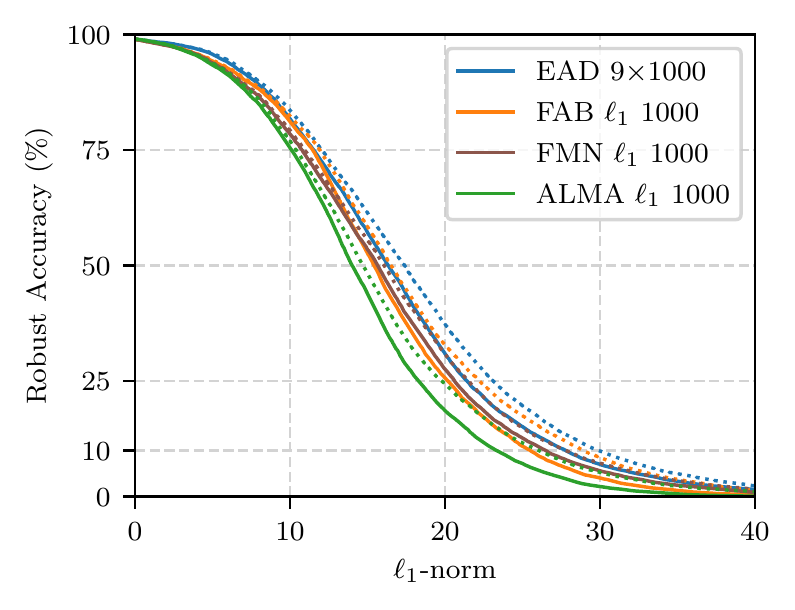}}\\
    \subfloat[SmallCNN-TRADES]{\includegraphics[width=0.7\columnwidth]{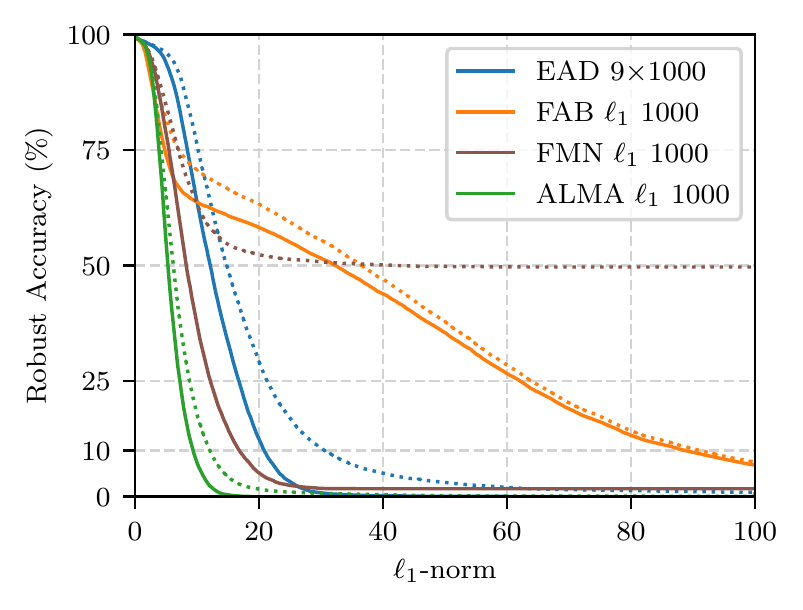}}\\
    \subfloat[CROWN-IBP]{\includegraphics[width=0.7\columnwidth]{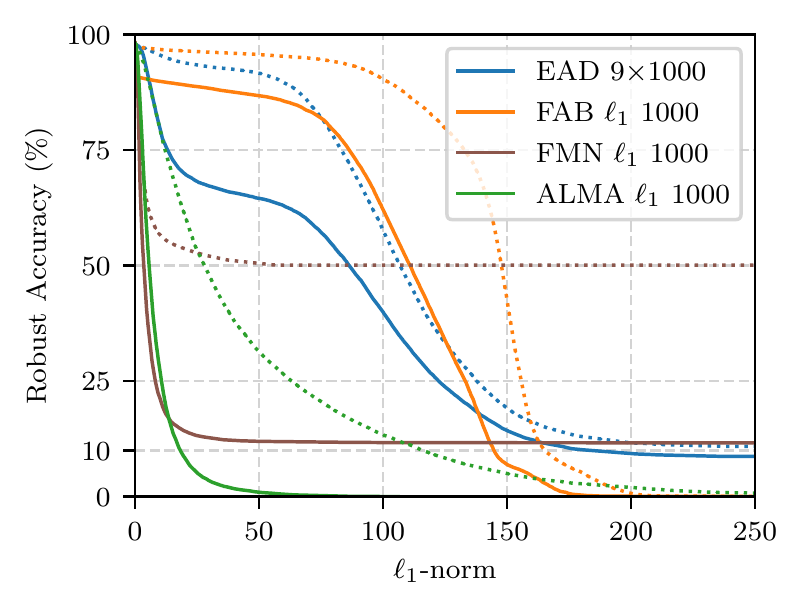}}
    \caption{Robust accuracy curves for MNIST models against $\ell_1$ attacks.}
    \label{fig:mnist_l1_curves}
\end{figure}

\begin{figure}
    \centering
    \subfloat[SmallCNN]{\includegraphics[width=0.7\columnwidth]{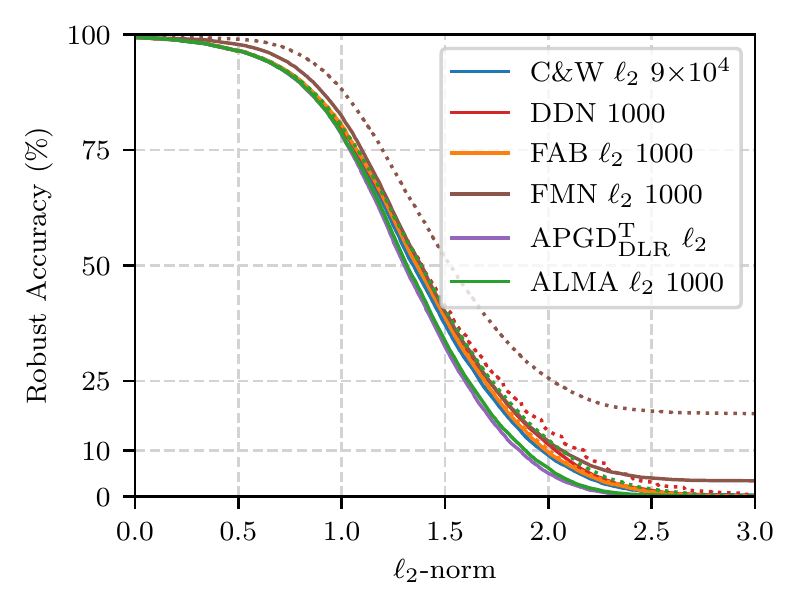}}\\
    \subfloat[SmallCNN-DDN]{\includegraphics[width=0.7\columnwidth]{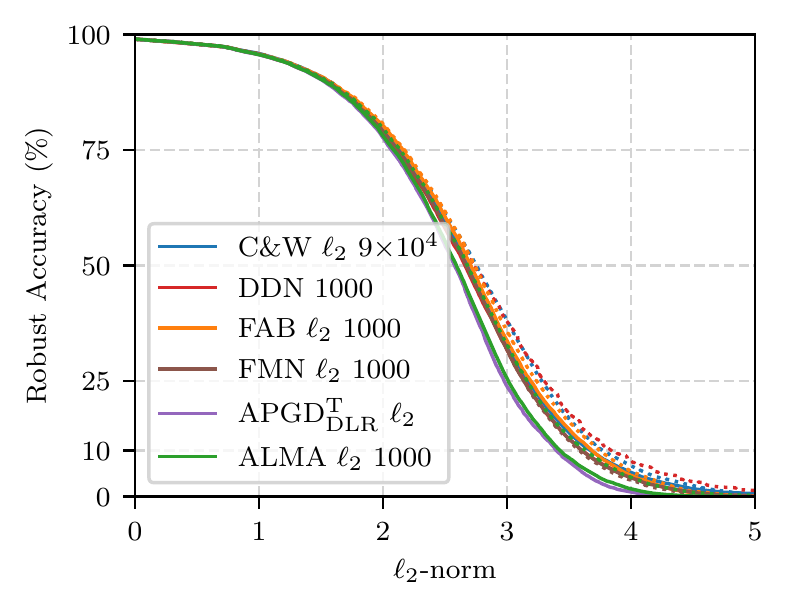}}\\
    \subfloat[SmallCNN-TRADES]{\includegraphics[width=0.7\columnwidth]{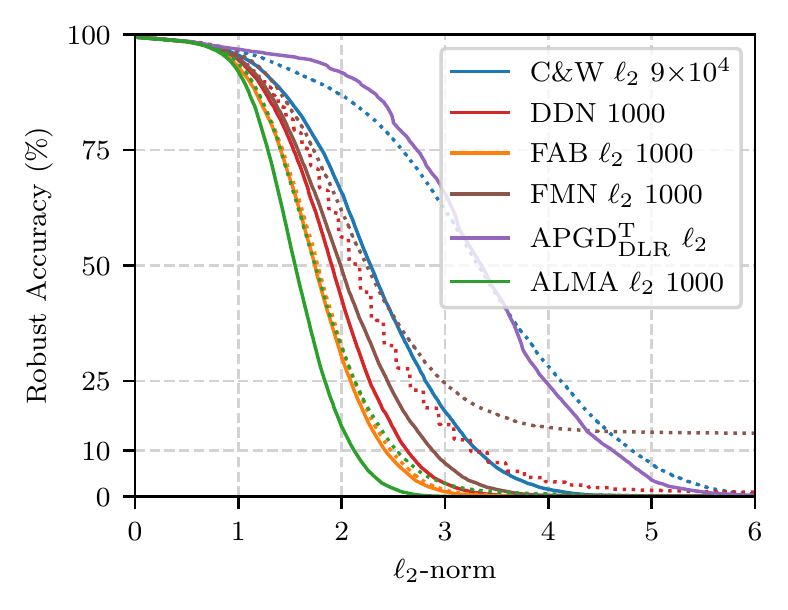}}\\
    \subfloat[CROWN-IBP]{\includegraphics[width=0.7\columnwidth]{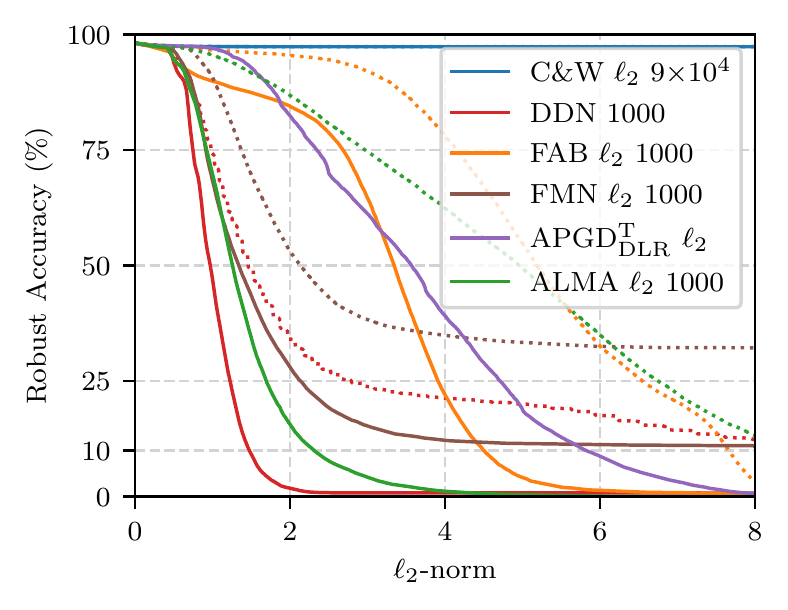}}\\
    \caption{Robust accuracy curves for MNIST models against $\ell_2$ attacks.}
    \label{fig:mnist_l2_curves}
\end{figure}

\begin{figure}
    \centering
    \subfloat[Wide ResNet 28-10]{\includegraphics[width=\columnwidth]{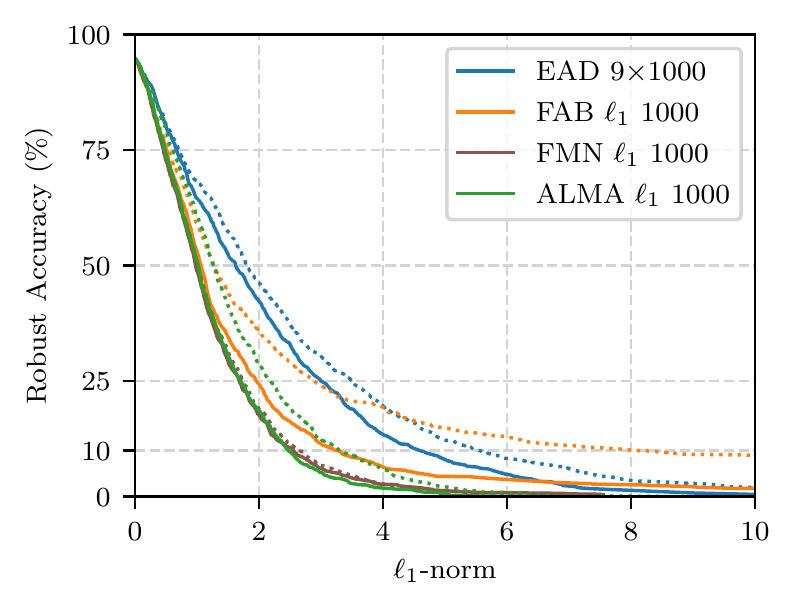}}\\
    \subfloat[Wide ResNet 28-10 Carmon \etal \cite{carmon2019unlabeled}]{\includegraphics[width=\columnwidth]{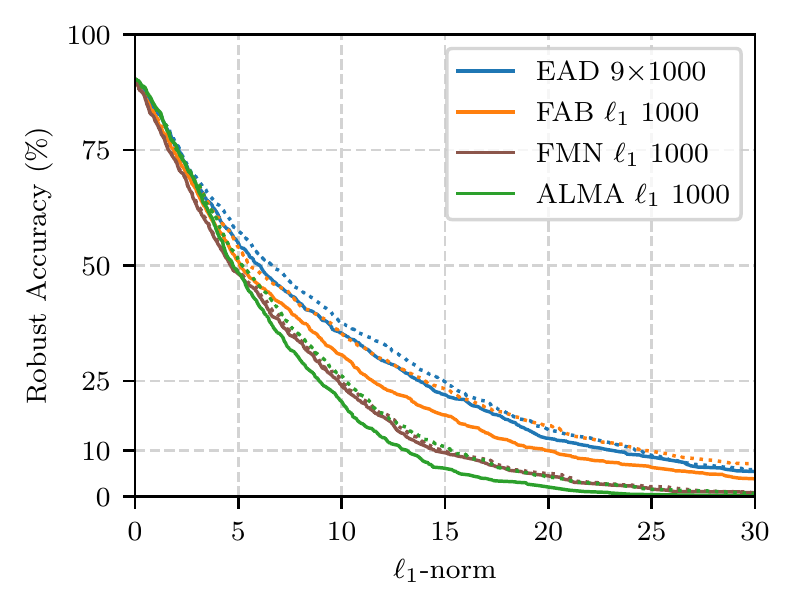}}\\
    \subfloat[ResNet-50 Augustin \etal \cite{augustin2020adversarial}]{\includegraphics[width=\columnwidth]{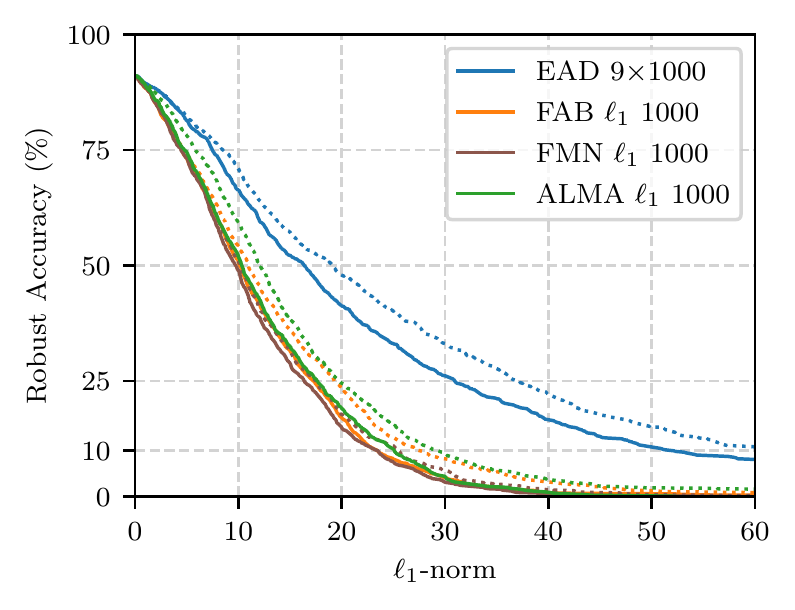}}
    \caption{Robust accuracy curves for CIFAR10 models against $\ell_1$ attacks.}
    \label{fig:cifar10_l1_curves}
\end{figure}

\begin{figure}
    \centering
    \subfloat[Wide ResNet 28-10]{\includegraphics[width=\columnwidth]{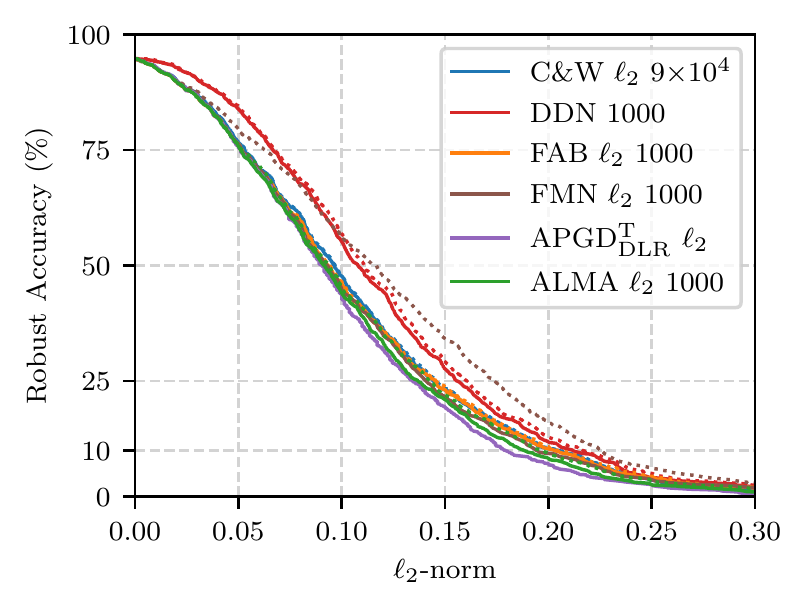}}\\
    \subfloat[Wide ResNet 28-10 Carmon \etal \cite{carmon2019unlabeled}]{\includegraphics[width=\columnwidth]{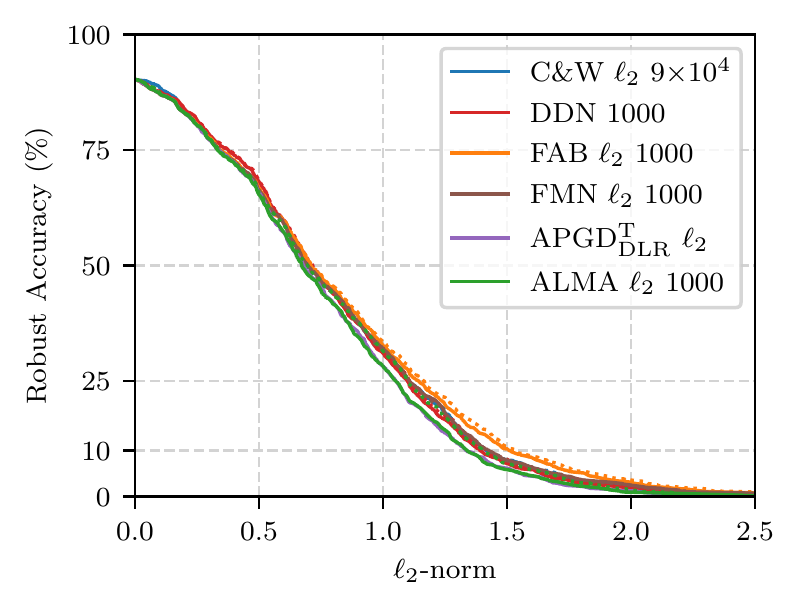}}\\
    \subfloat[ResNet-50 Augustin \etal \cite{augustin2020adversarial}]{\includegraphics[width=\columnwidth]{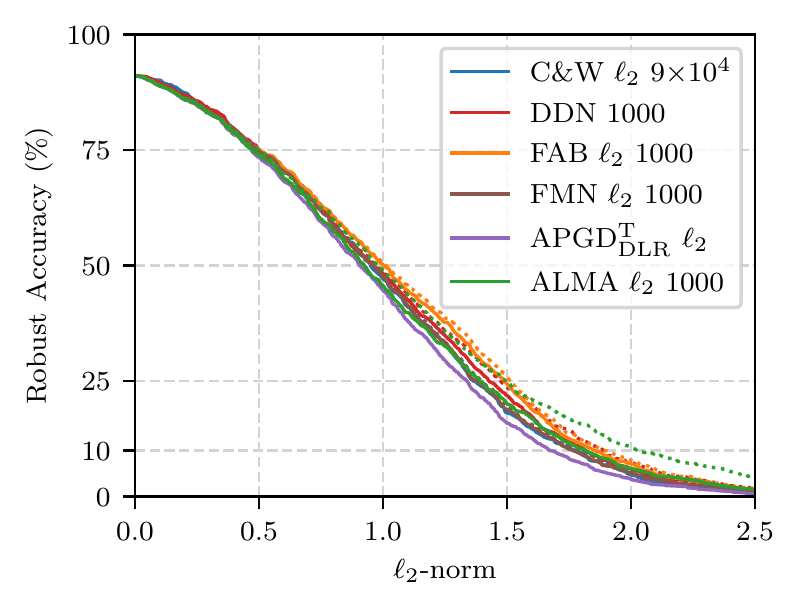}}
    \caption{Robust accuracy curves for CIFAR10 models against $\ell_2$ attacks.}
    \label{fig:cifar10_l2_curves}
\end{figure}

\begin{figure}
    \centering
    \subfloat[Wide ResNet 28-10]{\includegraphics[width=\columnwidth]{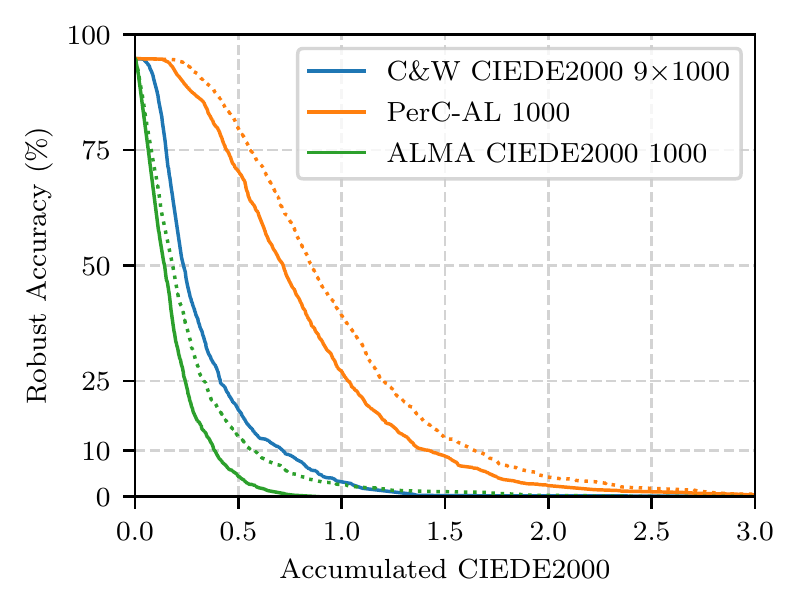}}\\
    \subfloat[Wide ResNet 28-10 Carmon \etal \cite{carmon2019unlabeled}]{\includegraphics[width=\columnwidth]{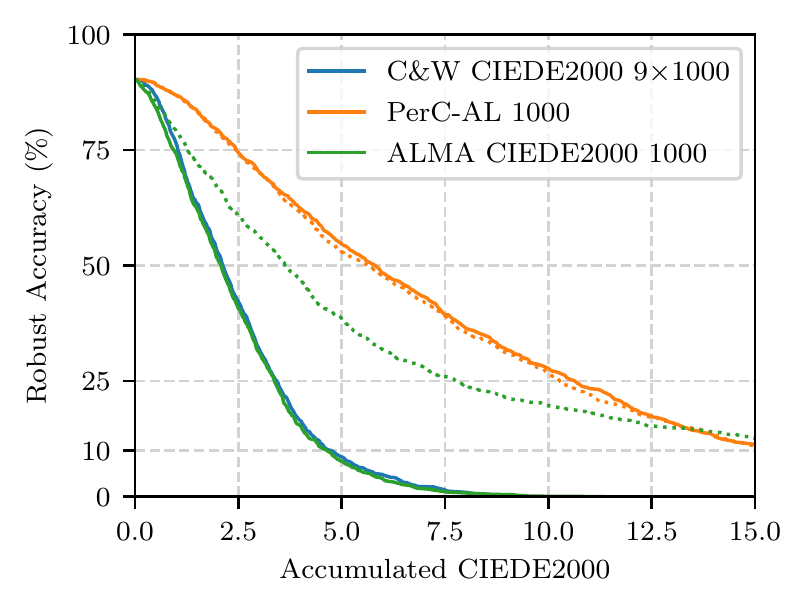}}\\
    \subfloat[ResNet-50 Augustin \etal \cite{augustin2020adversarial}]{\includegraphics[width=\columnwidth]{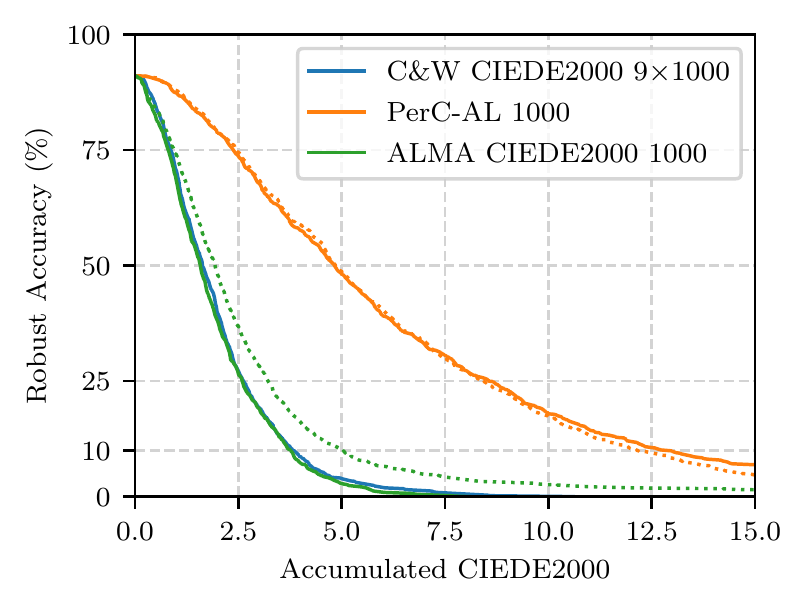}}
    \caption{Robust accuracy curves for CIFAR10 models against CIEDE2000 attacks.}
    \label{fig:cifar10_ciede2000_curves}
\end{figure}

\begin{figure}
    \centering
    \subfloat[ResNet-50]{\includegraphics[width=\columnwidth]{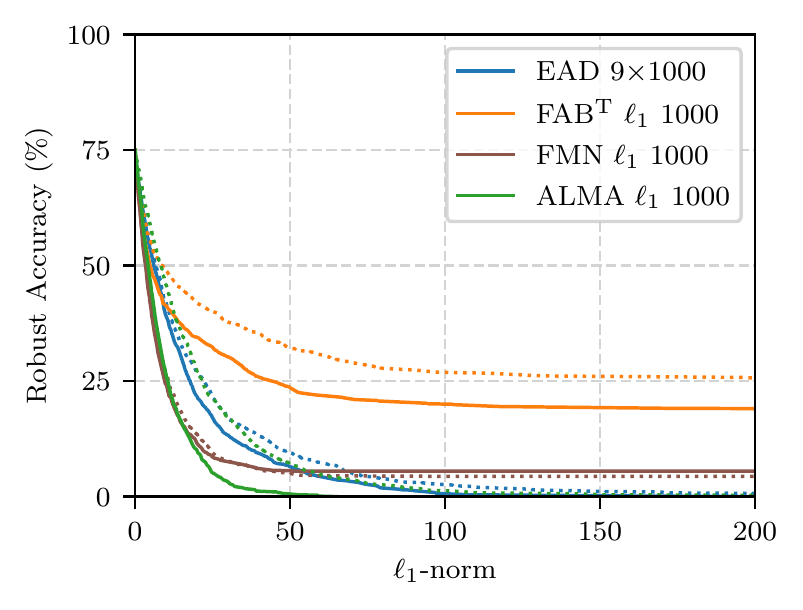}}\\
    \subfloat[ResNet-50 $\ell_2$ adv. trained]{\includegraphics[width=\columnwidth]{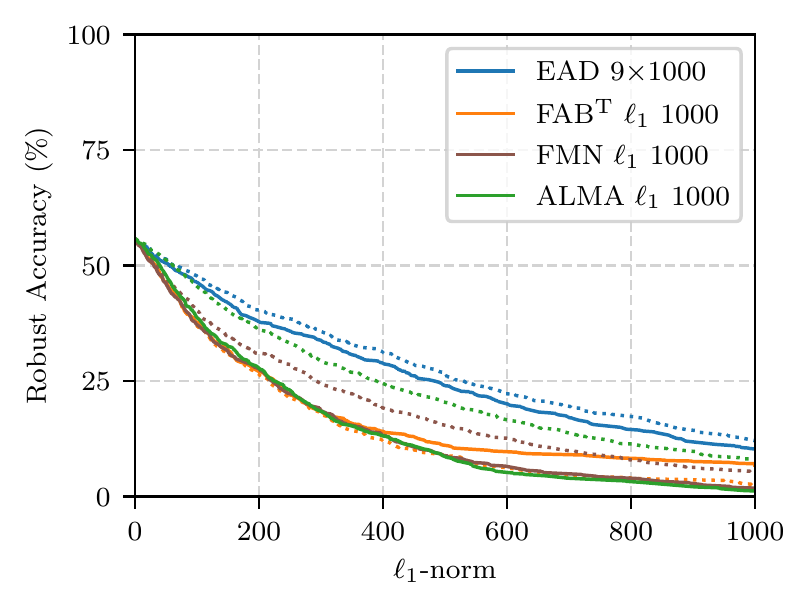}}\\
    \subfloat[ResNet-50 $\ell_\infty$ adv. trained]{\includegraphics[width=\columnwidth]{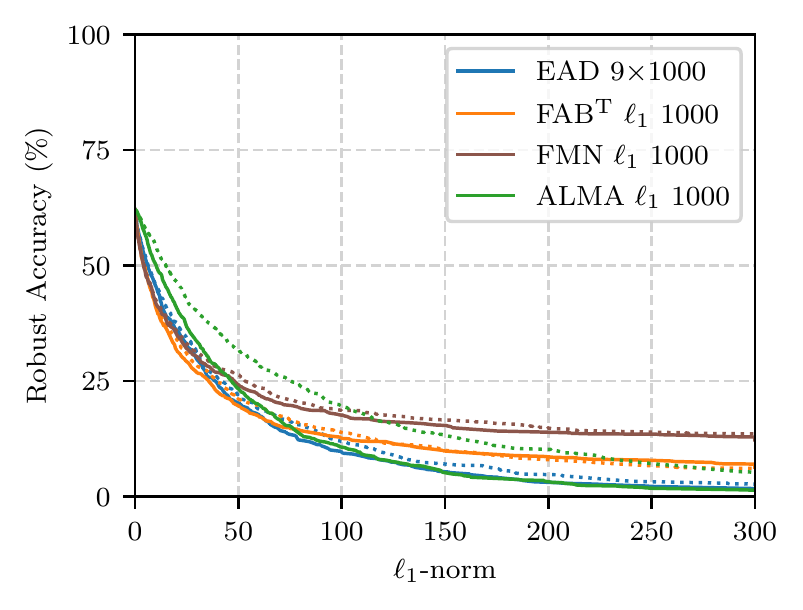}}
    \caption{Robust accuracy curves for ImageNet models against $\ell_1$ attacks.}
    \label{fig:imagenet_l1_curves}
\end{figure}

\begin{figure}
    \centering
    \subfloat[ResNet-50]{\includegraphics[width=\columnwidth]{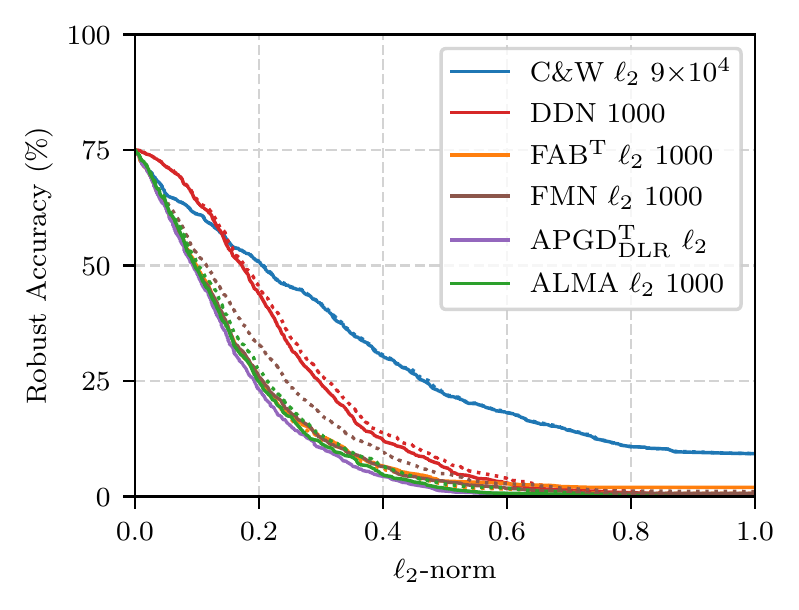}}\\
    \subfloat[ResNet-50 $\ell_2$ adv. trained]{\includegraphics[width=\columnwidth]{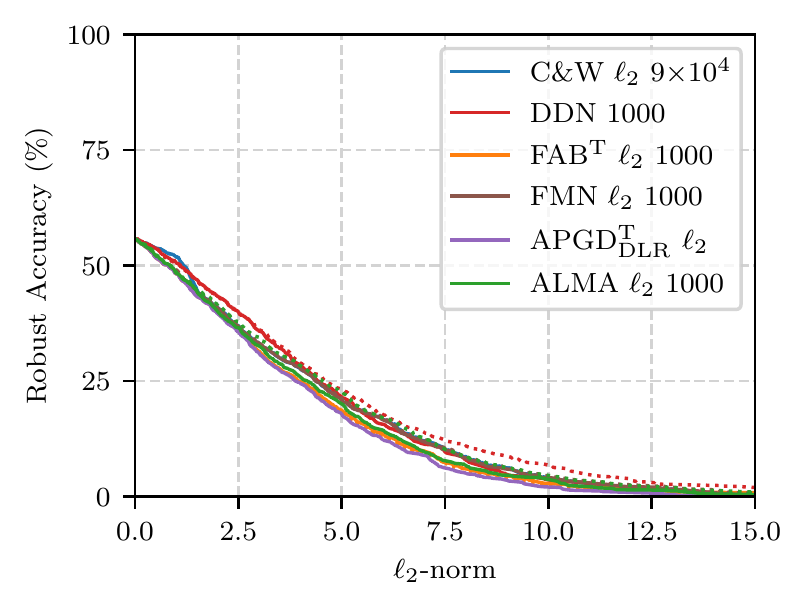}}\\
    \subfloat[ResNet-50 $\ell_\infty$ adv. trained]{\includegraphics[width=\columnwidth]{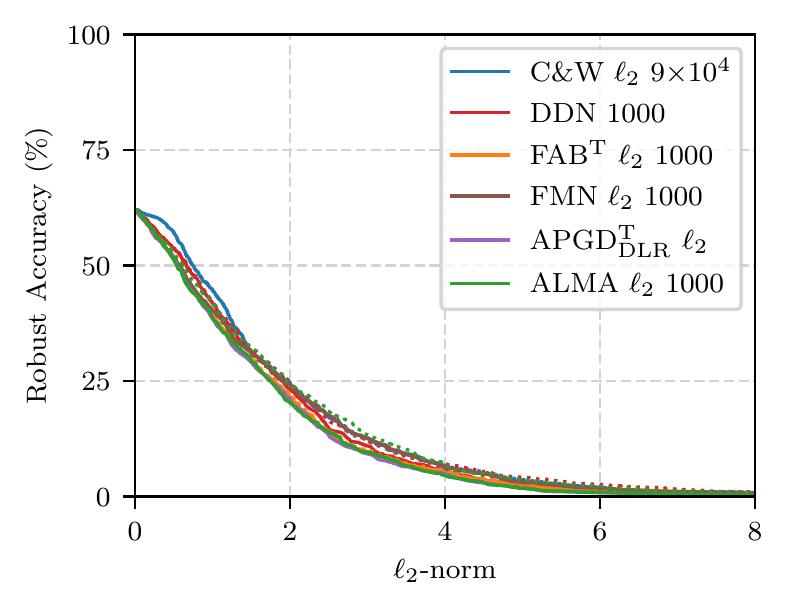}}
    \caption{Robust accuracy curves for ImageNet models against $\ell_2$ attacks.}
    \label{fig:imagenet_l2_curves}
\end{figure}

\begin{figure}
    \centering
    \subfloat[ResNet-50]{\includegraphics[width=\columnwidth]{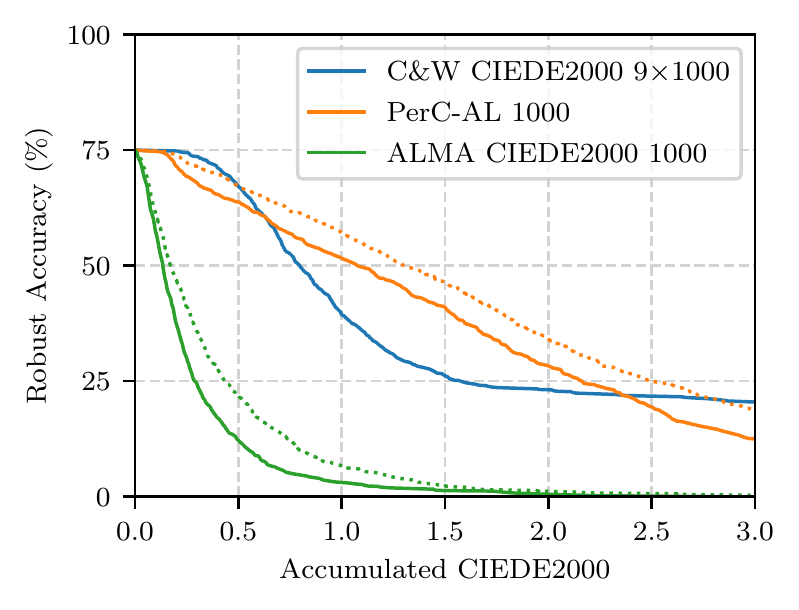}}\\
    \subfloat[ResNet-50 $\ell_2$ adv. trained]{\includegraphics[width=\columnwidth]{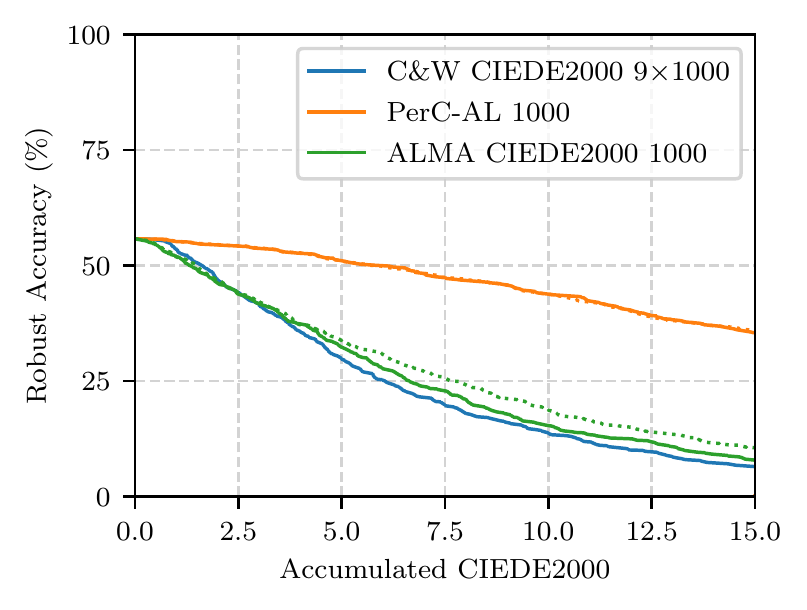}}\\
    \subfloat[ResNet-50 $\ell_\infty$ adv. trained]{\includegraphics[width=\columnwidth]{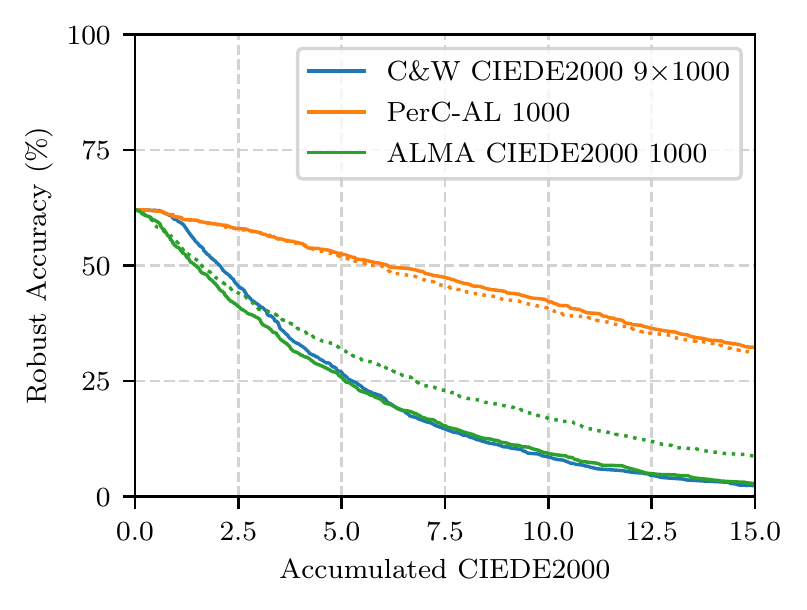}}
    \caption{Robust accuracy curves for ImageNet models against CIEDE2000 attacks.}
    \label{fig:imagenet_ciede2000_curves}
\end{figure}

\Autoref{fig:mnist_l1_curves,  fig:mnist_l2_curves, fig:cifar10_l1_curves, fig:cifar10_l2_curves, fig:cifar10_ciede2000_curves, fig:imagenet_l1_curves, fig:imagenet_l2_curves, fig:imagenet_ciede2000_curves} present the robust accuracy curves for each dataset and model against the attacks considered for each distance. The dotted line represent the reduced budget versions of the attack, as reported in the corresponding tables.

\begin{figure}
    \centering
    \subfloat[Wide ResNet 28-10]{\includegraphics[width=\columnwidth]{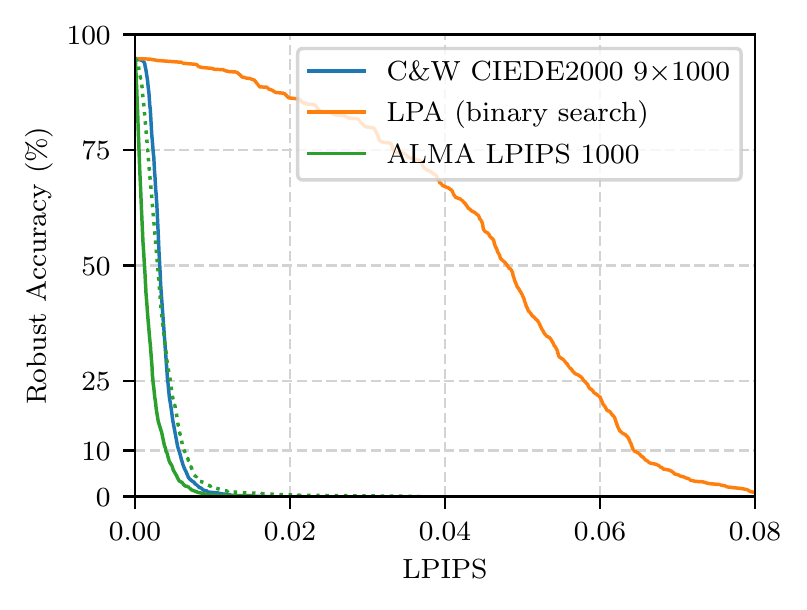}}\\
    \subfloat[Wide ResNet 28-10 Carmon \etal \cite{carmon2019unlabeled}]{\includegraphics[width=\columnwidth]{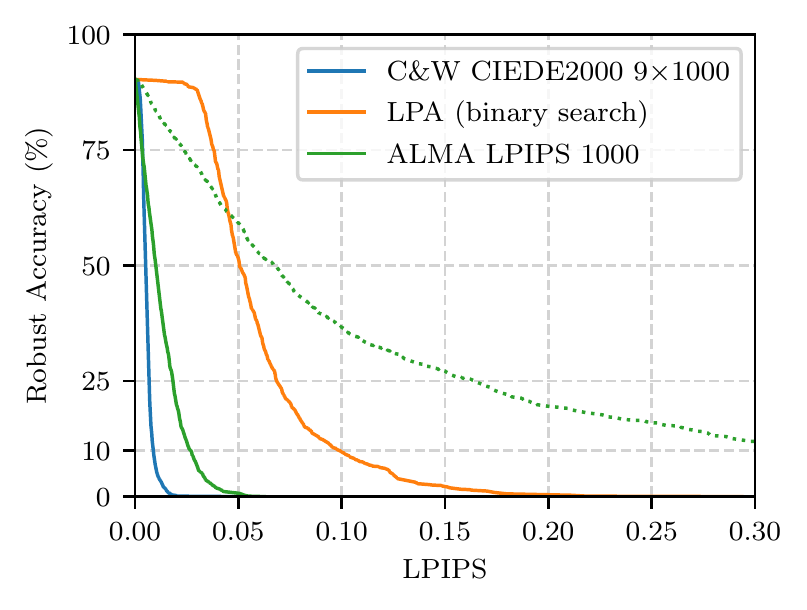}}\\
    \subfloat[ResNet-50 Augustin \etal \cite{augustin2020adversarial}]{\includegraphics[width=\columnwidth]{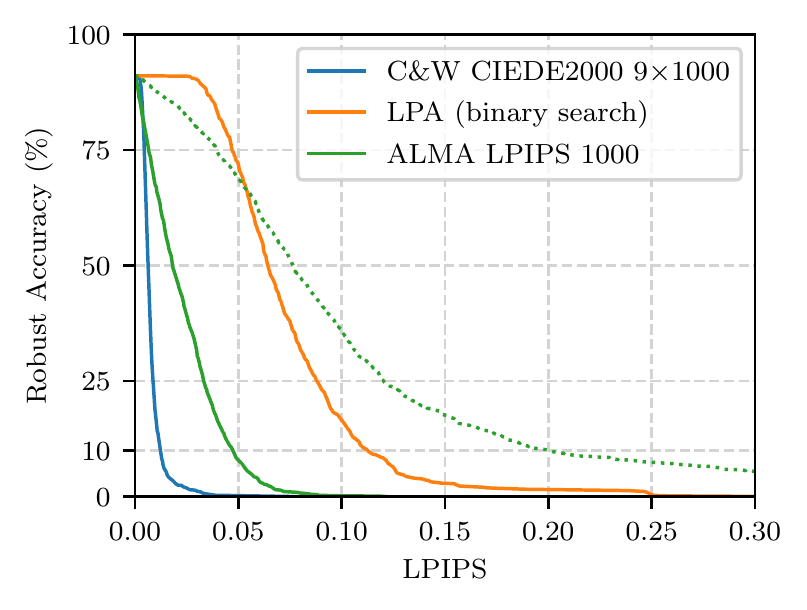}}
    \caption{Robust accuracy curves for CIFAR10 models against LPIPS attacks.}
    \label{fig:cifar10_lpips_curves}
\end{figure}

\begin{figure}
    \centering
    \subfloat[ResNet-50]{\includegraphics[width=\columnwidth]{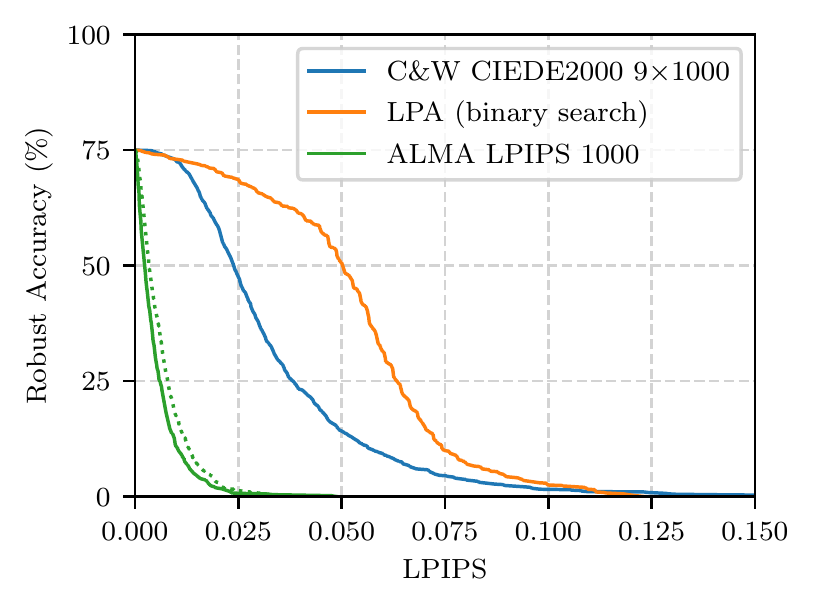}}\\
    \subfloat[ResNet-50 $\ell_2$ adv. trained]{\includegraphics[width=\columnwidth]{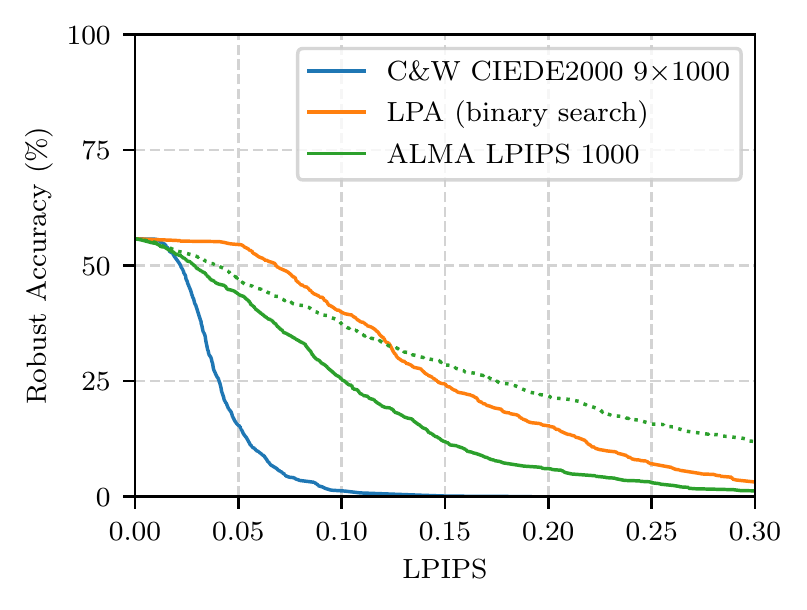}}\\
    \subfloat[ResNet-50 $\ell_\infty$ adv. trained]{\includegraphics[width=\columnwidth]{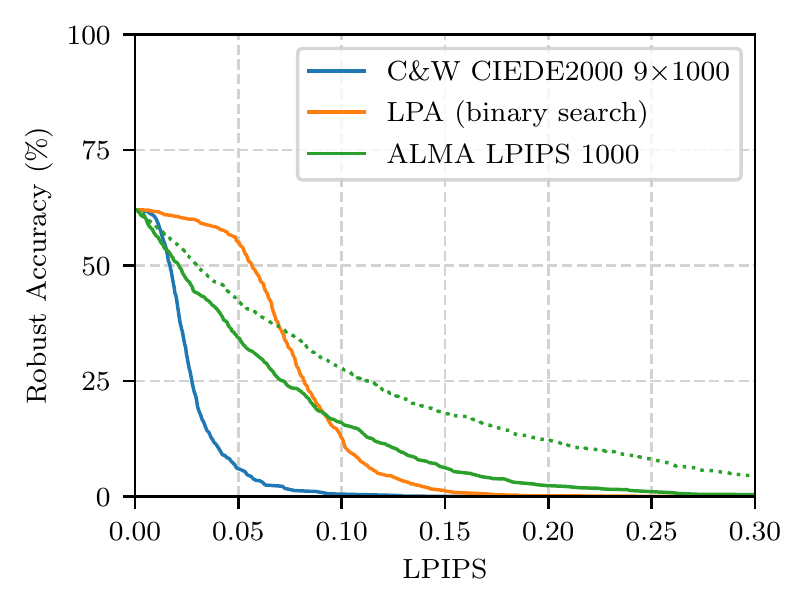}}
    \caption{Robust accuracy curves for ImageNet models against LPIPS attacks.}
    \label{fig:imagenet_lpips_curves}
\end{figure}

\end{document}